\documentclass[10pt,twocolumn,letterpaper]{article}
\usepackage[T1]{fontenc}

\usepackage{cvpr}              

\usepackage{amsmath,amsfonts,bm}

\def\eqref#1{equation~\ref{#1}}

\def\1{\bm{1}}

\def\vp{{\bm{p}}}
\def\vq{{\bm{q}}}

\DeclareMathAlphabet{\mathsfit}{\encodingdefault}{\sfdefault}{m}{sl}
\SetMathAlphabet{\mathsfit}{bold}{\encodingdefault}{\sfdefault}{bx}{n}

\newcommand{\method}[1]{\textsc{#1}}

\newcommand{\APBEV}{AP$_\text{BEV}$\xspace}
\newcommand{\AP}{AP$_\text{3D}$\xspace}

\DeclareRobustCommand{\eg}{e.g.\@\xspace}
\DeclareRobustCommand{\ie}{i.e.\@\xspace}
\def\vs{\emph{vs}\onedot}

\usepackage{graphicx}
\usepackage{amsmath}
\usepackage{amssymb}
\usepackage{booktabs}

\usepackage{multirow}
\usepackage{array}
\usepackage{url}
\usepackage{tablefootnote}
\usepackage{xspace}
\usepackage{algorithm}
\usepackage{algpseudocode}
\usepackage{float}
\usepackage{enumitem}
\usepackage[accsupp]{axessibility}  

\usepackage{color}
\usepackage[table]{xcolor}

\usepackage[pagebackref,breaklinks,colorlinks]{hyperref}
\definecolor{mydarkblue}{rgb}{0,0.08,0.45}
\definecolor{mydarkred}{rgb}{0.45,0.15,0}
\hypersetup{
  citecolor=mydarkblue,
  urlcolor=blue
}
\usepackage[capitalize]{cleveref}
\crefname{section}{Sec.}{Secs.}
\Crefname{section}{Section}{Sections}
\Crefname{table}{Table}{Tables}
\crefname{table}{Tab.}{Tabs.}

\def\cvprPaperID{****} 
\def\confName{CVPR}
\def\confYear{2022}

\newcommand{\lyft}{Lyft\xspace}
\newcommand{\nusc}{nuScenes\xspace}
\newcommand{\kitti}{KITTI\xspace}
\newcommand{\lidar}{LiDAR\xspace}

\newcommand{\ptc}{\boldsymbol{P}}

\newcommand{\cptc}{\boldsymbol{S}}

\newcommand{\pp}{PP\xspace}

\newcommand\mypara[1]{\vspace{1.mm}\noindent\textbf{#1}}

\newcommand{\ourmethod}{\method{MODEST}\xspace}

\makeatletter
\newcommand*{\@rowstyle}{}
\newcommand*{\rowstyle}[1]{
 \gdef\@rowstyle{#1}%
 \@rowstyle\ignorespaces%
}
\newcolumntype{=}{
>{\gdef\@rowstyle{}}%
}
\newcolumntype{+}{
>{\@rowstyle}%
}
\newcommand{\PreserveBackslash}[1]{\let\temp=\\#1\let\\=\temp}
\newcolumntype{C}[1]{>{\PreserveBackslash\centering}p{#1}}
\makeatother

\begin{document}

\title{Learning to Detect Mobile Objects
from LiDAR Scans Without Labels
}

\author{
Yurong You\thanks{Denotes equal contribution.} \thanks{Correspondences could be directed to \texttt{yy785@cornell.edu}} $^{,1}$\hspace{10pt}
Katie Luo\footnotemark[1] $^{,1}$\hspace{10pt}
Cheng Perng Phoo$^{1}$\hspace{10pt}
Wei-Lun Chao$^{2}$\\
Wen Sun$^{1}$\hspace{10pt}
Bharath Hariharan$^{1}$\hspace{10pt}
Mark Campbell$^{1}$\hspace{10pt}
Kilian Q. Weinberger$^{1}$\\ 
$^1$Cornell University\hspace{14pt}$^2$The Ohio State University
}
\maketitle

\begin{abstract}
Current 3D object detectors for autonomous driving are almost entirely trained on human-annotated data. Although of high quality, the generation of such data is laborious and costly, restricting them to a few specific locations and object types. 
This paper proposes an alternative approach entirely based on unlabeled data, which can be collected cheaply and in abundance almost everywhere on earth. 
Our approach leverages several simple common sense heuristics to create an initial set of approximate seed labels. For example, relevant traffic participants are generally not persistent across multiple traversals of the same route,  do not fly, and are never under ground. 
We demonstrate that these seed labels are highly effective to bootstrap a surprisingly accurate detector through repeated self-training without a single human annotated label. 
Code is available at \url{https://github.com/YurongYou/MODEST}.
\end{abstract}

\begin{figure*}[htb]
    \centering
    \includegraphics[width=\linewidth]{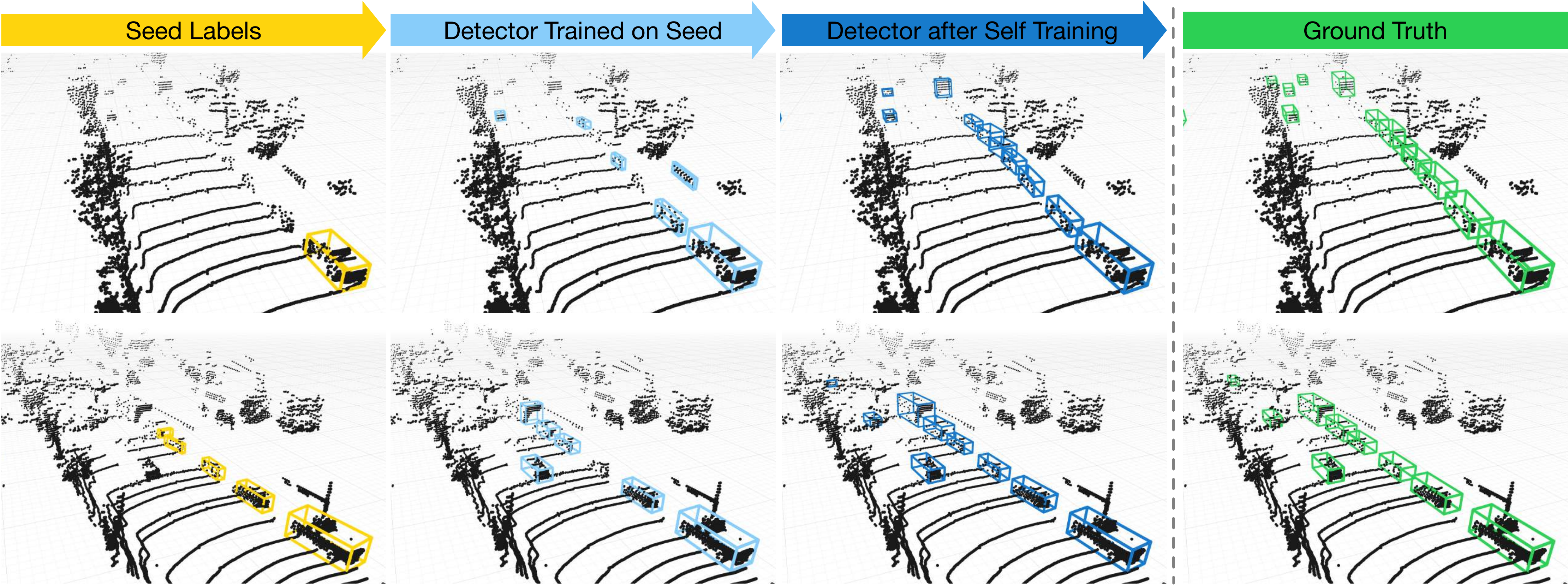}
    \vspace{-15pt}
    \caption{\textbf{Visualizations of \ourmethod outputs.} We show \lidar scans from two scenes in the \lyft dataset in two rows. From \textit{zero labels}, our method is able to bootstrap a detector that achieves results close to the ground truth. 
    The key insight is to utilize noisy ``seed" labels produced from an ephemerality score and filtered with common-sense properties, and self-train upon them to obtain high quality results.
    \label{fig:teaser}
    }
    \vspace{-15pt}
\end{figure*}

\section{Introduction}
\label{sec:intro}

Autonomous driving promises to revolutionize how we transport goods, travel, and interact with our environment. To safely plan a route, a self-driving vehicle must first perceive and localize mobile traffic participants such as other vehicles and pedestrians in 3D. Current state-of-the-art 3D object detectors are all based on deep neural networks~\cite{qi2018frustum,shi2019pointrcnn,yang2018pixor,shi2020pv} and  can yield up to $80$ average precision on benchmark datasets\cite{geiger2012we,geiger2013vision}.

However, as with all deep learning approaches, these techniques have an insatiable need for labeled-data.
Specifically, to train a 3D object detector that takes LiDAR scans as input, one typically needs to first come up with a list of objects of interest and annotate each of them with tight bounding boxes in the 3D point cloud space. Such a data annotation process is laborious and costly, but worst of all, the resulting detectors only achieve high accuracy when the training and test data distributions match~\cite{wang2020train}. In other words, their accuracy deteriorates over time and space, as looks and shapes of cars, vegetation, and background objects change.  
To guarantee good performance, one has to collect labeled training data for specific geo-fenced areas and re-label data constantly, greatly limiting the applicability and development of  self-driving vehicles.

These problems motivate the question: \emph{Can we learn a 3D object detector from unlabeled LiDAR data?} 
Here, we focus on ``mobile'' objects, \ie, objects that might move, which cover a wide range of traffic participants.
At first glance, this seems insurmountably challenging. After all, 
how could a detector know just from the LiDAR point cloud that a pedestrian is a traffic participant and a tree is not? We tackle this problem with the help of two important insights: 
1) we can use simple heuristics that, even without labeling, can occasionally distinguish traffic participants from background objects more or less reliably; 2) if data is noisy but diverse, neural networks excel at identifying the common patterns, allowing us to repeatedly self-label the remaining objects, starting from a small set of seed labels. 

\mypara{Weak labels through heuristics.} We build upon a simple yet highly generalizable concept to discover mobile objects
--- \emph{mobile objects are unlikely to stay persistent at the same location over time.}
While this requires unlabeled data at multiple timestamps for the same locations, collecting them is arguably cheaper than annotating them.
After all, many of us drive through the same routes every day (\eg, to and from work or school). Even when going to new places, the new routes for us are likely frequent for the local residents.

Concretely, whenever we discover multiple traversals of one route, we calculate  a simple ephemerality statistic \cite{barnesephemerality} for each LiDAR point, which characterizes the change of its local neighborhood across traversals.
We cluster LiDAR points according to their coordinates and ephemerality statistics. Resulting clusters with high ephemerality statistics,  and located on the ground, are considered as mobile objects and are further fitted with upright bounding boxes. 
 
\mypara{Self-training (ST).}
While this initial \emph{seed} set of mobile objects is not exhaustive (\eg, parked cars may be missed) and somewhat noisy in shape, we demonstrate that an object detector trained upon them can already learn the underlying object patterns and is able to output more and higher-quality bounding boxes than the seed set.
This intriguing observation further opens up the possibility of using the detected object boxes as ``better'' pseudo-ground truths to train a new object detector.
We show that such a self-training cycle~\cite{lee2013pseudo, xie2020self} enables the detector to improve itself over time; notably, it can even benefit from additional, unlabeled data that do not have multiple past traversals associated to them.

We validate our approach, \ourmethod (\textbf{M}obile \textbf{O}bject \textbf{D}etection with \textbf{E}phemerality and \textbf{S}elf-\textbf{T}raining) on the Lyft Level 5 Perception Dataset~\cite{lyft2019} and the \nusc Dataset~\cite{caesar2020nuscenes} with various types of detectors~\cite{shi2019pointrcnn, lang2019pointpillars, zhou2018voxelnet, yan2018second}. We demonstrate that \ourmethod
yields remarkably accurate mobile object detectors, comparable to their supervised counterparts.
Concretely, our contributions are three-fold:
\begin{enumerate}[noitemsep,topsep=2pt] 
\item We propose a simple, yet effective approach to identifying ``seed" mobile objects from multiple traversals of LiDAR scans using \textit{zero labels}.
\item We show that using these seed objects,
we can bootstrap accurate mobile object detectors via self-training.
\item We evaluate our method exhaustively under various setting and demonstrate consistent performance across multiple real-world datasets.
\end{enumerate}
\section{Related Works}
\label{sec:related}
We seek to build object detectors without any human supervision.
We briefly discuss several related research areas. 

\mypara{3D object detection and existing datasets.} Most existing 3D object detectors take 3D point clouds generated by LiDAR as input. They either consist of specialized neural architectures that can operate on point clouds directly \cite{qi2017pointnet, qi2017pointnet++, qi2018frustum, shi2019pointrcnn, yang20203dssd} or voxelize the point clouds to leverage 2D or 3D convolutional neural architectures \cite{zhou2018voxelnet, zhou2019end, yang2018pixor, yan2018second, shi2020pv, liang2020learning, lang2019pointpillars, chen2017multi}. Regardless of the architectures, they are trained using supervision and their performances hinges directly on the training dataset. However, the limited variety of objects and driving conditions in existing autonomous driving datasets \cite{geiger2012we, geiger2013vision, lyft2019, caesar2020nuscenes, Argoverse} impedes the generalizability of the resulting detectors~\cite{wang2020train}. 

Generalizing these to new domains requires a fresh labeling effort.
In contrast, our unsupervised approach automatically discovers all the traffic participants, and can be used to train detectors in any new condition without any labeling.

\mypara{Unsupervised Object Discovery in 2D/3D.} Our work follows prior work on discovering objects both from 2D images as well as from 3D data. 
A first step in object discovery is to identify candidate objects, or ``proposals'' from a single scene/image.
For 2D images, this is typically done by segmenting the image using appearance cues\cite{ma2015simultaneous, garcia2015saliency, vo2019unsupervised, cho2015}, but color variations and perspective effects make this difficult.
Tian \etal~\cite{tian2021unsupervised} exploits the correspondence between images and 3D point clouds to detect objects in 2D.
In 3D scenes, one can use 3D information such as surface normals~\cite{shin2010unsupervised, triebel2010segmentation, garcia2015saliency, kochanov2016scene, herbst2011toward, herbst2011rgb, karpathy2013object, collet2011structure}.
One can also use temporal changes such as motion~\cite{du2020unsupervised, ma2014unsupervised, choudhary2014slam, ma2015simultaneous, kochanov2016scene}.
Our work combines effective 3D information with cues from changes in the scene over time to detect mobile objects\cite{herbst2011toward, herbst2011rgb, mason2012object}. In particular, similar to our approach, Herbst et al.\cite{herbst2011toward, herbst2011rgb} reconstruct the same scene at various times and carve out dissimilar regions as mobile objects. 
We use the analogous idea of ephemerality as proposed by Barnes et al.\cite{barnesephemerality}.
We show in our work that this idea yields a surprisingly accurate set of initial objects.
In addition, we also leverage other common-sense rules such as locations of the objects (\eg objects should stay on the ground)  \cite{chen20173d, mason2014unsupervised, collet2015herbdisc, collet2013exploiting} or shapes of an object (\eg objects should be compact) \cite{karpathy2013object, collet2015herbdisc, collet2013exploiting}.
However, crucially, we do not just stop at this proposal stage.
Instead, we use these seed labels to train an object detector through multiple rounds of self-training.
This effectively identifies objects consistent across multiple scenes.
While previous work has attempted to use this consistency cue\cite{vo2019unsupervised, cho2015, vo2020toward, shin2010unsupervised, triebel2010segmentation, zhang2013unsupervised, abbeloos20173d} (including co-segmentation\cite{faktor2013co, yuan2017deep, li2018deep}), prior work typically uses clustering to accomplish this.
In contrast, we demonstrate that neural network training and self-training provides a very strong signal and substantially improves the quality of the proposals or seed labels.

\mypara{Self-training, semi-supervised and self-supervised learning.} 
When training our detector, we use self-training, which has been shown to be highly effective for semi-supervised learning\cite{lee2013pseudo, xie2020self}, domain adaptation \cite{zou2018unsupervised, chen2019progressive, zou2019confidence, zhang2019category, mei2020instance, yang2021st3d} and few-shot/transfer learning \cite{tian2020rethinking, phoo2021STARTUP, Phoo_2021_ICCV, Ghiasi_2021_multi_ST}. Interestingly, we show that self-training can not only discover more objects, but also correct the initially noisy box labels. 
This result that neural networks can denoise noisy labels has been observed before~\cite{han2018co,pleiss2020identifying,arpit2017closer, Pathak2017Learning}.  
Self-training also bears resemblance to other semi-supervised learning techniques\cite{grandvalet2005entropy,  berthelot2019mixmatch,berthelot2019remixmatch,sohn2020fixmatch} but is simpler and more broadly applicable.

\section{Method}
\label{sec:method}
\mypara{Problem setup.} 
We want a detector that detects \emph{mobile} objects, \ie, objects that might move, in LiDAR point clouds.
We wish to train this detector only from \emph{unlabeled} data obtained simply by driving around town, using a car equipped with synchronized sensors (in particular, LiDAR which provides 3D point clouds and GPS/INS which provides accurate estimates of vehicle position and orientation).
Such a data collection scheme is practical and requires no annotators; indeed, it can be easily collected as people go about their daily lives.
We assume that this unlabeled data include at least a few locations that have been visited \emph{multiple times}; as we shall see, this provides us with a very potent learning signal for identifying mobile objects.

\mypara{Overview.} We propose simple, high-level common-sense properties that can easily identify a few \emph{seed} objects in the unlabeled data.
These discovered objects then serve as labels to train an off-the-shelf object detector. 
Specifically, building upon the neural network's ability to learn consistent patterns from initial seed labels, we bootstrap the detector
by self-training \cite{lee2013pseudo, xie2020self} using the same unlabeled data. The self-training process serves to correct and expand the initial pool of seed objects, gradually discovering more and more objects to further help train the detector. The whole process is summarized in \autoref{alg:main}. 

\begin{figure*}[t]
    \centering
    \includegraphics[width=\linewidth]{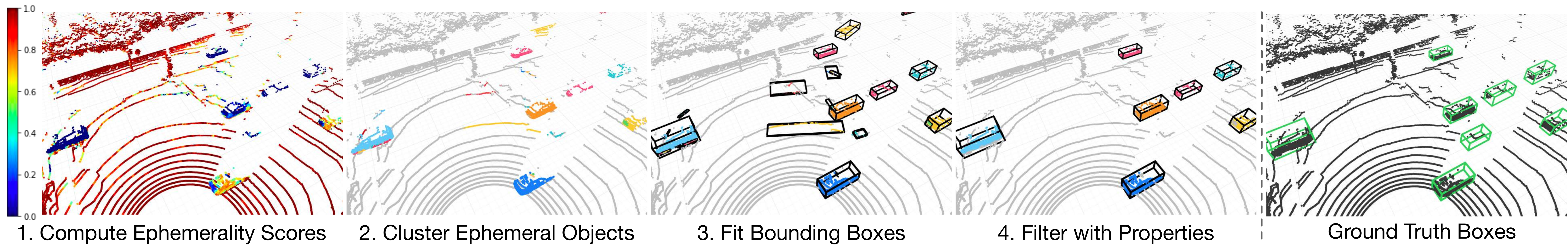}
    \vspace{-15pt}
    \caption{\textbf{Generation of seed labels.} Seed labels from object discovery are used to train downstream detectors. We begin by computing the \pp score for each point. Then we segment out clusters that are non-persistent and apply a box-fitting algorithm to each cluster. We filter out superfluous bounding boxes using our common-sense assumptions. This entire process is supervision-free. 
    \label{fig:seed_generation}}
    \vspace{-10pt}
\end{figure*}

\subsection{Discovering objects through common-sense}
What properties define mobile objects or traffic participants?
Clearly, the most important characteristic is that they are \emph{mobile}, \ie, they move around.
If such an object is spotted at a particular location (\eg, a car at an intersection), it is unlikely that the object will still be there when one visits the intersection again a few days hence.
In other words, mobile objects are \emph{ephemeral} members of a scene~\cite{barnesephemerality}.
Of course, occasionally mobile objects like cars might be parked on the road for extended periods of time.
However, for the most part ephemeral objects are likely to be mobile objects.

What other properties do mobile objects have? 
It is clear that they must be on the ground, not under the ground or above in the sky.
They are also likely to be smaller than buildings.
One can come up with more, but we find that these intuitive, common-sense properties serve as sufficient constraints for mining objects from unlabeled data.

Building upon these two sets of properties, we propose a bottom-up approach, which begins with identifying points that are ephemeral, followed by clustering them into seed objects that obey these common-sense properties. In the following sections, we discuss the implementations and visualize an example seed label generation in \autoref{fig:seed_generation}.

\vspace{-5pt}
\subsubsection{Identifying ephemeral points}
We assume that our unlabeled data include a set of locations $L$ which are traversed multiple times in separate driving sessions (or \emph{traversals}).
For every traversal $t$ through location $c \in L$, we aggregate point clouds captured within a range of $[-H_s, H_e]$ of $c$ to produce a dense 3D point cloud $\cptc_c^t$ for location $c$ in traversal $t$\footnote{We can easily transform captured point clouds to a shared coordinate frame via precise localization information through GPS/INS.}.
We then use these dense point clouds $\cptc_c^t$ to define ephemerality as described by Barnes et al.~\cite{barnesephemerality}.
Concretely, to check if a 3D point $\vq$ in a scene is ephemeral,
for each traversal $t$ we can count the number $N_t(\vq)$ of \lidar points that fall within a distance $r$ to $\vq$, 
\begin{equation}
        N_t \left(\boldsymbol{q}\right) = \left|\left\{\boldsymbol{p}_i \mid \|\boldsymbol{p}_i - \boldsymbol{q}\|_{2} < r, \boldsymbol{p}_i \in \cptc_c^t \right\}\right|.
\end{equation}
{If $\vq$ is part of the static background, then its local neighborhood will look the same in all traversals, and so the counts $N_t(\vq)$ will be all similar.}
Thus, we can check if $\vq$ is ephemeral by checking if $N_t(\vq)$ is approximately uniform across traversals.
To this end, we compute
\begin{equation}
    P \left(t ;\boldsymbol{q}\right) = 
    \frac{N_t \left(\boldsymbol{q}\right)} {\sum_{t'=1}^T N_{t'} \left(\boldsymbol{q}\right)}.\label{eq:p}
\end{equation}
and define the \emph{persistence point score} (PP score) as:
\begin{align}
\tau(\vq) = \left\{ \begin{array}{ll} 0 & \textrm{ if } N_t(\vq) = 0 \;\;\forall t; \\  \frac{H\left(P(t;\vq)\right)}{\log(T)} & \textrm{ otherwise. } \end{array} \right. 
\end{align}
Here $H(\cdot)$ is the information entropy,   $T$ is the number of traversals through location $c$, and $\log{(T)}$ a normalizer to guarantee that $\tau(\mathbf{q})\in[0,1]$.\footnote{The KL divergence between $P$ and the uniform distribution $f(t) = 1/T$ is $KL(P(t;\vq)|| f(t)) =\log(T)-H\left(P(t;\vq)\right)$ and high entropy implies large similarity with the uniform distribution. }
A high PP score implies a high entropy of the distribution $P(\cdot; \vq)$, which means that the counts $N_t(\vq)$ over $t$ are all similar, indicating that the point $\vq$ is part of the static background (see 1. in \autoref{fig:seed_generation}).

\subsubsection{From ephemeral points to ephemeral objects}\label{ss_ephemeral_obj}
We compute the \pp score for each point in the LiDAR point clouds collected at a location $c$ using multiple traversals.
This automatically surfaces non-persistent (and thus mobile) objects as blobs of points with \emph{low \pp scores}. 
We segment out these blobs automatically using the following straightforward clustering approach. 
First, we construct a graph whose vertices are points in the point cloud.
The edges in the graph connect each point only to its mutual $K$-nearest neighbors in 3D within a distance $r'$.
Each edge between points $\vp$ and $\vq$ is assigned a weight equal to the difference between their \pp scores, \ie,
\begin{equation}
\label{eq::edge_weight}
    w(e_{\vq, \vp}) = |\tau(\vq) - \tau(\vp)|.
\end{equation}
The graph structure together with the edge weights define a new metric that quantifies the similarity between two points. In this graph, two points that are connected by a path are considered to be close if the path has low total edge weight, namely, the points along the path share similar \pp scores, indicating these points are likely from the same object. In contrast, a path in the graph that has high total edge weight likely goes across the boundary of two different objects (e.g., a mobile object and the background). 
Many graph-based clustering algorithms can fit the bill.
We deploy the widely used DBSCAN~\cite{ester1996density} algorithm for the clustering due to its simplicity and its ability to cluster without the need of setting the number of clusters beforehand. 
DBSCAN returns a list of clusters, from which we remove clusters of static (and thus persistent) background points by 
applying a threshold $\gamma$ on the $\alpha$ percentile of \pp scores in a cluster (i.e., remove the cluster if the $\alpha$ percentile of the \pp scores in this cluster is larger than $\gamma$). 
We then apply a straightforward bounding box fitting algorithm \cite{zhang2017efficient} to fit an up-right bounding box to each cluster.

\vspace{-5pt}
\subsubsection{Filtering using other common sense properties}

Finally, following our common-sense assumptions, we remove bounding boxes that are under the ground plane, floating in the air, or having exceptional large volume (see the supplementary). 
This produces the final set of \emph{seed} pseudo-ground-truth bounding boxes for mobile objects.

\mypara{Discussion.} Our proposed procedure for object discovery, while intuitive and fully unsupervised, has several limitations. First, the bounding boxes are produced only for locations that were traversed multiple times.
Second, owing to many filtering steps that we apply, these bounding boxes are not exhaustive.
For example, parked cars along the side of the road might be marked as persistent and therefore may not be identified as mobile objects.
Finally, our heuristic box-fitting approach might fit inaccurate bounding boxes to noisy clusters that contain background points or miss foreground points.
Thus in general, this initial set of seed bounding boxes might (a) miss many objects, and (b) produce incorrect box shapes and poses. 
Nevertheless, we find that this seed set has enough signal for training a high quality mobile object detector, as we discuss below.

{
\setlength{\textfloatsep}{0pt}
\begin{algorithm}[t]
\caption{Mobile Object Detection with Ephemerality and Self-Training (\ourmethod) \label{alg:main}}
\begin{algorithmic}
\Require $\{\ptc_i\}$ \lidar{}s with accurate localization, \\
\hspace{20pt} $I_{\max}$ maximum self-training (ST) iterations
\Ensure $D_{I_{\max}}$ the mobile detector after $I_{\max}$ rounds ST
\vskip 5pt
\State $\{\{S^t\}_i\} \gets$ aggregate \lidar from other traversals
\State $\{\tau(\ptc_i)\} \gets \{\text{compute\_\pp}(\{S^t\}_i,\ptc_i)\}$
\State $\mathcal{B}_0 \gets \{\text{cluster\_fit\_boxes\_filter}(\tau(\ptc_i), \ptc_i)\}$
\State $D_0 \gets \text{train\_detector}(\{\ptc_i\}, \mathcal{B}_0)$ \Comment{$0$-th round training}
\For{$j \gets 1$ to $I_{\max}$} \Comment{$j$-th round self-training}
    \State $\mathcal{B}_j \gets \text{get\_detection}(D_{j-1}, \{\ptc_i\})$ \Comment{pseudo-labels}
    \State $\mathcal{B}_j \gets \text{filter\_by\_\pp}(\mathcal{B}_j, \{\ptc_i, \tau(\ptc_i)\})$
    \State $D_j \gets \text{train\_detector}(\{\ptc_i\}, \mathcal{B}_j)$
\EndFor
\end{algorithmic}
\end{algorithm}
}
\subsection{Bootstrapping a mobile object detector}

Concretely, we simply take off-the-shelf 3D object detectors~\cite{shi2019pointrcnn,lang2019pointpillars,zhou2018voxelnet,yan2018second} and directly train them \emph{from scratch} on these initial seed labels via minimizing the corresponding detection loss from the detection algorithms.

Intriguingly, the object detector trained in this way \emph{outperforms} the original seed bounding boxes themselves --- \emph{the ``detected'' boxes have higher recall and are more accurate than the ``discovered'' boxes on the same training point clouds.} See \autoref{fig:teaser} for an illustration.
This phenomenon of a neural network improving on the provided noisy labels themselves is superficially surprising, but it has been observed before in other contexts~\cite{Pathak2017Learning}.
The key reason is that the noise is not consistent: the initial labels are generated scene-by-scene and object-by-object. In some scenes a car may be missed because it was parked throughout all traversals, while in many others it will be discovered as a mobile object. Even among discovered boxes of similar objects, some may miss a few foreground points but others wrongly include background points.
The neural network, equipped with limited capacity\footnote{We note that for detection problems, a neural network can hardly over-fit the training data to achieve $100\%$ accuracy~\cite{pan2021model}, in contrast to classification problems~\cite{zhang2021understanding}.}, thus cannot reproduce this inconsistency and instead instead identifies the underlying consistent object patterns. 

In particular, we find that the detector substantially improves recall: it is able to identify objects (like parked cars) that were missed in the ephemerality computation, because these seemingly static objects are nevertheless similar in shape to other moving objects identified as ephemeral.
We also find many cases where the detector automatically corrects the box shape, based on the average box shape of similar objects it has encountered in the training data.
Finally, because the detector no longer needs multiple traversals, it can also find new mobile objects in scenes that were only visited once.
In summary, this initial detector already discovers far more objects than the initial seed set, and localizes them more accurately.

\mypara{Automatic improvement through self-training.}
Given that the trained detector has discovered many more objects, we can use the detector itself to produce an improved set of ground-truth labels, and re-train a new detector from scratch with these better ground truths.
Furthermore, we can \emph{iterate} this process: the new retrained detector has more positives and more accurate boxes for training, so it will likely have higher recall and better localization than the initial detector.
As such we can use this second detector to produce a new set of pseudo-ground-truth boxes which can be used to train a third detector and so on.
This iterative self-training~\cite{lee2013pseudo, xie2020self} process will eventually converge when the pseudo-ground truth labeling is consistent with itself and the detector can no longer improve upon it. 

While it is possible for this iterative self-training to cause concept drift (\eg, the detector reinforces from its error), 
we find empirically that a simple thresholding step similar to that in \autoref{ss_ephemeral_obj} --- remove the pseudo-ground-truth box if the $\alpha$ percentile of the \pp scores within the box is larger than $\gamma$ 
--- is highly effective in removing false positives (hence improve precision) in the self-training process and prevents performance degradation (\autoref{fig:ablation_rounds}).

\section{Experiments}
\label{sec:experiments}
\mypara{Datasets.} We validate our approach on two datasets: Lyft Level 5 Perception~\cite{lyft2019} and nuScenes~\cite{caesar2020nuscenes}. 
To the best of our knowledge, these are the only two publicly available autonomous driving datasets that have both bounding box annotations and multiple traversals with accurate localization. To ensure fair assessment of generalizability, we re-split the dataset so that the training set and test set are \emph{geographically disjoint}; we also discard locations with less than 2 examples in the training set. This results a train/test split of 11,873/4,901 point clouds for \lyft and 3,985/2,324 for \nusc. To construct ground truth labels, we group all the traffic participants types in the original datasets into a single ``mobile" object. Note that \emph{the ground-truth labels are only used for evaluation, not training. }

In addition, we convert the raw \lyft and \nusc data into the \kitti format to leverage off-the-shelf 3D object detectors that is predominantly built for \kitti~\cite{geiger2013vision}. We use the roof LiDAR (40 or 60 beam in \lyft; 32 beam in \nusc), and the global 6-DoF localization along with the calibration matrices directly from the raw data. 

\mypara{On localization.} With current localization technology, we can reliably achieve accurate localization (\eg, \emph{1-2\,cm-level} accuracy with RTK\footnote{\url{https://en.wikipedia.org/wiki/Real-time_kinematic_positioning}}, \emph{10\,cm-level} with Monte Carlo Localization scheme~\cite{chong2013synthetic} as adopted in the \nusc dataset~\cite{caesar2020nuscenes}). We assume good localization in the training set.

\mypara{Evaluation metric.} We follow \kitti~\cite{geiger2012we} to evaluate object detection in the bird's-eye view (BEV) and in 3D for the mobile objects. We report average precision (AP) with the intersection over union (IoU) thresholds at 0.5/0.25, which are used to evaluate cyclists and pedestrians objects in \kitti. 
We further follow \cite{wang2020train} to evaluate the AP at various depth ranges.
Due to space constraints, we only present evaluation results with IoU=0.25 in \Cref{tbl:main,tbl:ablation,tbl:nusc_025,tbl:label_quality,tbl:kitti_025,tbl:recall_by_class_iou025_summary}. Please refer to the supplementary for results with IoU=0.5.

\mypara{Implementation. } We present results on PointRCNN~\cite{shi2019pointrcnn} (the conclusions hold for other detectors such as PointPillars~\cite{lang2019pointpillars}, and VoxelNet (SECOND)~\cite{zhou2018voxelnet, yan2018second}. See more details in the supplementary materials). For reproducibility, we use the publicly available code from OpenPCDet~\cite{openpcdet2020} for all models. We use the default hyperparameters tuned for KITTI except on the \lyft dataset in which we enlarge the perception range from 70m to 90m (since Lyft provides labels beyond 70m) and reduce the number of training epochs by 1/4 (since the training set is about three times of the size of \kitti). We default to 10 rounds of self-training (chosen arbitrarily due to compute constraints) and trained the model from scratch for each round of self-training. We also include results on PointRCNN trained up to 40 rounds, where we empirically observe that the performance converges (\autoref{fig:ablation_rounds}). All models are trained with 4 NVIDIA 3090 GPUs. Please refer to the supplementary materials for full hyperparameters.

\mypara{Baselines and ablations.} We are the first work to train object detectors without any labels at all, and as such no previously published baselines exist. We create baselines by ablating the two key components of our model: seed labels generation via multiple traversals and repeated self-training: 
\begin{itemize}[noitemsep,topsep=1pt]
    \item \ourmethod-\pp(R0): This is a detector trained with seed labels generated \emph{without} leveraging the multiple traversals. The seed labels are constructed by the exact same process as described in \autoref{sec:method}, except we replace the edge weights in \autoref{eq::edge_weight} by spatial proximity: $w(e_{\vq, \vp}) = \|\vq - \vp\|_2$ and change $\epsilon$ to $1.0$ in DBSCAN, and do not perform any \pp-score-based filtering on the clusters generated by DBSCAN. No repeated self-training is performed for this baseline. 
    \item \ourmethod-\pp(R$i$): This detector is trained similarly to the previous baseline except we repeat $i$ rounds of self-training witout using \pp-score-based filtering.
    \item \ourmethod(R0): This detector is trained with the seed labels without repeated self-training.
\end{itemize}

\begin{table}[!t]
\centering
\renewcommand{\arraystretch}{1.05}
\tabcolsep 2pt
\caption{\textbf{Detection performance with different methods on the \lyft dataset.} 
We report \APBEV / \AP with IoU=$0.25$ for mobile objects under various ranges.
R$i$ stands for $i$-th round self-training (R0 is training from seed labels). 
We also report the performance of detectors trained with ground-truth labels on the \kitti and the \lyft datasets at the last two rows. 
\vspace{-8pt}
\label{tbl:main}}
\resizebox{.47\textwidth}{!}{%
\begin{tabular}{=l|+C{44pt}|+C{44pt}|+C{44pt}||+C{44pt}}
\multicolumn{1}{c|}{\multirow{2}{*}{Method}} & \multicolumn{4}{c}{\APBEV / \AP @ IoU = 0.25}\\
\cline{2-5}\
 & 0-30 & 30-50 & 50-80 & 0-80 \\\hline
 \ourmethod-\pp (R0)
 & 46.4 / 45.4 & 16.5 / 10.8 & \phantom{0}0.9 / \phantom{0}0.4 & 21.8 / 18.0
\\
 \ourmethod-\pp (R10)
 & 49.9 / 49.3 & 32.3 / 27.0 & \phantom{0}3.5 /\phantom{0} 1.4 & 30.9 / 27.3 
 \\
 \ourmethod (R0)
 & 65.7 / 63.0 & 41.4 / 36.0 & \phantom{0}8.9 / \phantom{0}5.7 & 42.5 / 37.9 
 \\
 \ourmethod (R10)
 & 73.8 / 71.3 & 62.8 / 60.3 & 27.0 / 24.8 & 57.3 / 55.1 
 \\
 \ourmethod (R40)
 & 76.4 / 74.1 & 64.2 / 62.9 & 47.1 / 45.5 & 64.4 / 62.7
 \\
\hline\hline
\rowstyle{\leavevmode\color{gray}} Sup. (\kitti) 
& 79.3 / 78.9 & 57.2 / 56.6 & 30.8 / 29.8 & 58.6 / 57.3 
\\
\rowstyle{\leavevmode\color{gray}} Sup. (\lyft)
& 82.8 / 82.6 & 70.8 / 70.3 & 50.2 / 49.6 & 69.5 / 69.1 
\\
 \hline
\end{tabular}
}
\vspace{-8pt}
\end{table}

\begin{table}[!t]
\centering
\renewcommand{\arraystretch}{1.05}
\small
\tabcolsep 2pt
\caption{Detection results on the \nusc Dataset.
We report \APBEV / \AP at IoU=$0.25$ for mobile objects under various ranges. Please refer to \autoref{tbl:main} for naming.
\label{tbl:nusc_025}}
\vspace{-8pt}
\resizebox{.47\textwidth}{!}{%
\begin{tabular}{=l|+C{37pt}|+C{37pt}|+C{37pt}||+C{37pt}}
\multicolumn{1}{c|}{\multirow{2}{*}{Method}} & \multicolumn{4}{c}{\APBEV / \AP @IoU = 0.25}\\
\cline{2-5}\
 & 0-30 & 30-50 & 50-80 & 0-80 \\\hline
 \ourmethod-PP(R0)
 & \phantom{0}0.7\,/\,\phantom{0}0.1 & \phantom{0}0.0\,/\,\phantom{0}0.0 & \phantom{0}0.0\,/\,\phantom{0}0.0 & \phantom{0}0.2\,/\,\phantom{0}0.1\\
 
 \ourmethod-PP(R10) 
 & - & - & - & -\\
 
\ourmethod(R0) & 16.5\,/\,12.5 & \phantom{0}1.3\,/\,\phantom{0}0.8 & \phantom{0}0.3\,/\,\phantom{0}0.1 & \phantom{0}7.0\,/\,\phantom{0}5.0\\

 \ourmethod(R10) & 24.8\,/\,17.1 & \phantom{0}5.5\,/\,\phantom{0}1.4 & \phantom{0}1.5\,/\,\phantom{0}0.3 & 11.8\,/\,\phantom{0}6.6 \\\hline\hline
 
 \rowstyle{\leavevmode\color{gray}} Sup. (\nusc) & 39.8\,/\,34.5 & 12.9\,/\,10.0 & \phantom{0}4.4\,/\,\phantom{0}2.9 & 22.2\,/\,18.2 \\
 \hline
\end{tabular}%
}
\vspace{-15pt}
\end{table}

\subsection{Detecting mobile objects without annotations}
We present results on \lyft in \autoref{tbl:main} and observe that: 
\begin{enumerate}[noitemsep,topsep=1pt]
    \item \textbf{Object detectors can be trained using unlabeled data}: We observe that our approach yields accurate detectors, especially for the 0-50m range.
    For this range, \ourmethod is competitive with the fully supervised model trained on Lyft, and in fact \emph{outperforms} a model trained with ground-truth supervision on KITTI.
    This suggests that \ourmethod is especially useful for bootstrapping recognition models in new domains.
    \item \textbf{Our initial seed labels suffices to train a detector}: Detectors learned from our initial seed (\ourmethod(R0)) achieve more than 50\% of supervised performance for nearby objects, suggesting that our common sense cues do produce a good initial training set. We investigate the quality of the seed labels in \autoref{tbl:label_quality}. 
    \item \textbf{Repeated self-training significantly improves performance.} We observe that if the seed labels can provide enough signals for training a decent detector, self-training can further drastically boost the performance, for example, by more than 500\% from 8.9 to 47.1 on \APBEV IoU=0.25 on 50-80m range.
    \item \textbf{Ephemerality is a strong training signal.} We observe \ourmethod unanimously outperform \ourmethod-PP by a significant margin. 
\end{enumerate}

We note that this performance of our detectors is especially good considering that we are evaluating it on predicting the \emph{full amodal extent of the bounding box}, even though it has only seen the visible extent of the objects in the data.
We notice that because of this discrepancy, our model produces smaller boxes; a size adjustment might well substantially improve accuracy for higher overlap thresholds.

We apply our approach in \nusc dataset without changing the hyperparameters and report the results in \autoref{tbl:nusc_025}.
The above conclusions still hold. 
Notice that when we remove ephemerality (\ourmethod-PP(R$10$)), we are not able to extract enough signals to train a decent detector on \nusc where \lidar is much sparser than \lyft.

\begin{table}[!t]
\centering
\renewcommand{\arraystretch}{1.05}
\small
\tabcolsep 2pt
\caption{Detection performance on the \kitti validation set with models trained on the \lyft dataset.
We report \APBEV\,/\,\AP with IoU=$0.25$ for mobile objects under various ranges. Please refer to \autoref{tbl:main} for naming. 
\label{tbl:kitti_025}}
\vspace{-8pt}
\resizebox{.47\textwidth}{!}{%
\begin{tabular}{=l|+C{36pt}|+C{36pt}|+C{36pt}||+C{36pt}}
\multicolumn{1}{c|}{\multirow{2}{*}{Method}} & \multicolumn{4}{c}{\APBEV\,/\,\AP @ IoU = 0.25}\\
\cline{2-5}\
 & 0-30 & 30-50 & 50-80 & 0-80 \\\hline
 \ourmethod-PP (R10) 
 & 56.6\,/\,55.0 & 22.7\,/\,18.1 & \phantom{0}0.9\,/\,\phantom{0}0.7 & 42.7\,/\,40.7\\
 \ourmethod (R10)
 & 73.5\,/\,71.6 & 50.3\,/\,48.3 & \phantom{0}9.6\,/\,\phantom{0}8.1 & 61.7\,/\,59.7 \\
 \ourmethod (R40)
 & 73.6\,/\,73.2 & 49.9\,/\,48.2 & 15.1\,/\,13.9 & 63.0\,/\,61.1\\
 \hline\hline
 \rowstyle{\leavevmode\color{gray}} Sup. (\lyft) & 82.0\,/\,81.9 & 53.4\,/\,51.8 & 24.9\,/\,22.2 & 71.3\,/\,69.6 \\
 \rowstyle{\leavevmode\color{gray}} Sup. (\kitti) & 88.0\,/\,87.9 & 73.6\,/\,72.0 & 46.7\,/\,45.4 & 81.1\,/\,81.0 \\
 \hline
\end{tabular}%
}
\vspace{-15pt}
\end{table}

\mypara{Cross-domain evaluation.} 
It is possible that our automatic labeling process and multiple rounds of self-training overfit to biases in the training domain.
To see if this is the case we test whether our models trained on unlabeled data from \lyft generalize to \kitti (\autoref{tbl:kitti_025}). 
We observe that our detectors are still just as competitive with supervised detectors especially on close to middle ranges.

\subsection{Analysis}
Given the good performance of our detectors, we dig deeper into the individual components to identify the key contributors to success.

\begin{figure}
    \centering
    \includegraphics[width=\linewidth]{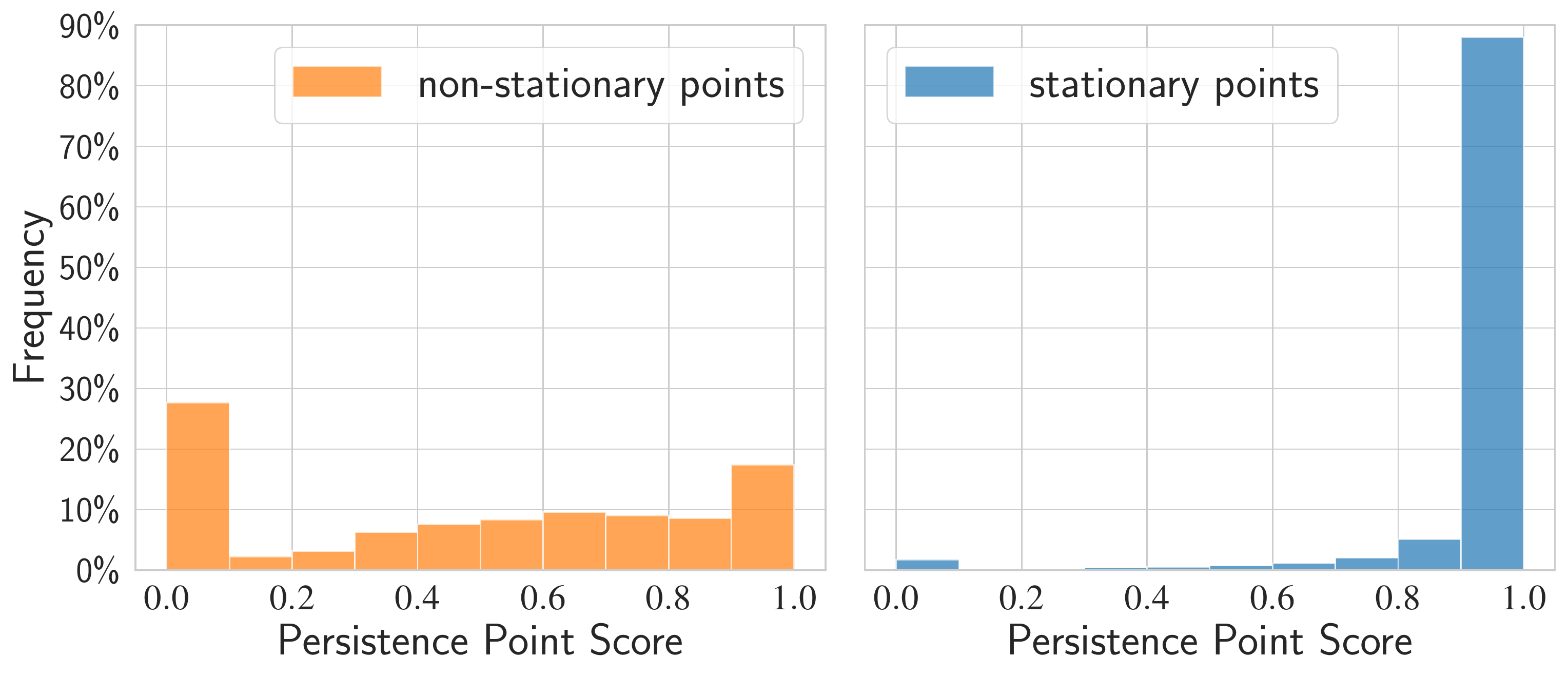}
    \vspace{-17pt}
    \caption{\textbf{Histogram of \pp score for non-stationary and stationary points.} We separate the non-stationary and stationary points by the ground-truth object labels, and compute corresponding \pp score frequency in the \lyft train set. 
    }
    \label{fig:pp_score}
    \vspace{-5pt}
\end{figure}
\mypara{Analysis on the \pp score.}
In \ourmethod, \pp score plays a critical role of distinguishing stationary points from stationary, background points. 
As such, in \autoref{fig:pp_score}, we plot the histogram of \pp score for non-stationary and stationary points in the train split of the Lyft dataset. Non-stationary points are defined as the points within labeled bounding boxes for mobile objects, while background points are the points outside of these bounding boxes. The histogram clearly shows that for background points the \pp score highly concentrates around $1$, while for non-stationary points the score is much lower. 

\begin{figure}
    \centering
    \begin{subfigure}{0.48\linewidth}
    \centering
    \includegraphics[width=\textwidth]{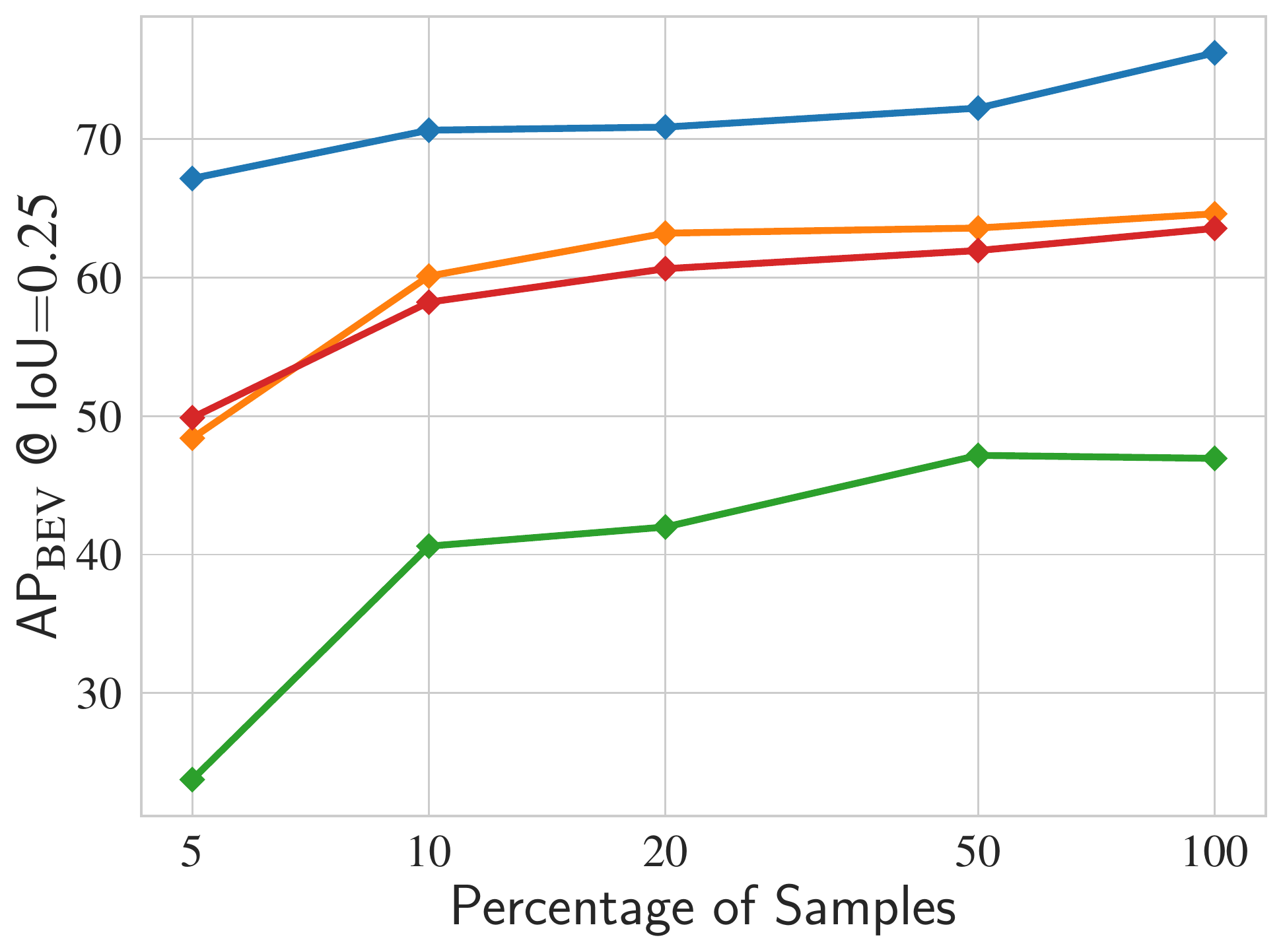}
    \caption{IoU=0.25 \label{fig:ablation_smaples_bev025}}
    \end{subfigure}
    \begin{subfigure}{0.48\linewidth}
    \centering
    \includegraphics[width=\textwidth]{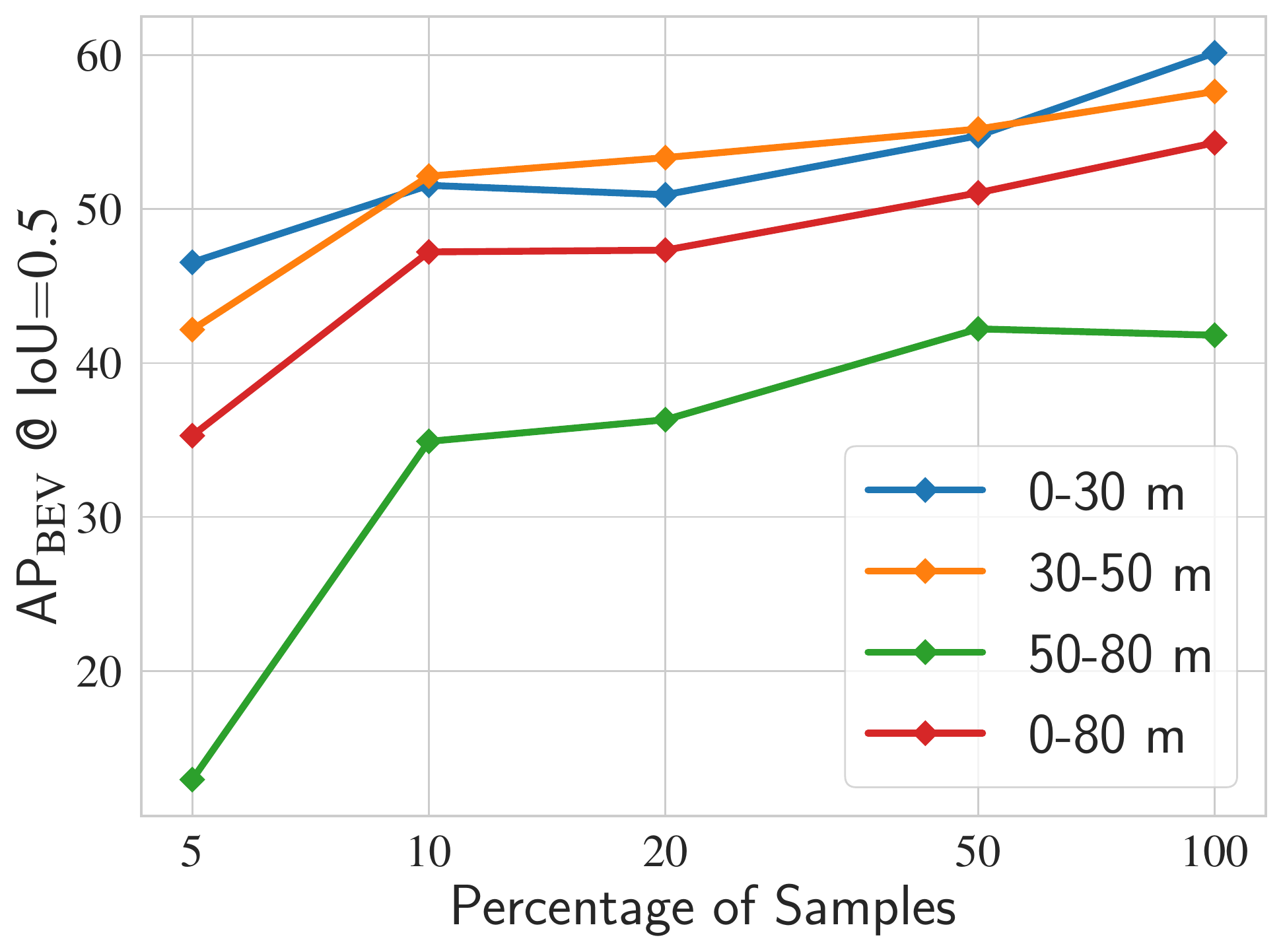}
    \caption{IoU=0.5 \label{fig:ablation_smaples_bev05}}
    \end{subfigure}
    \vspace{-8pt}
    \caption{\textbf{Number of ``train'' samples \vs the performance.} We report \APBEV with IoU=0.5 and IoU=0.25 for mobile objects in various ranges on \lyft test set from models trained with different mount of unlabeled data with 40 rounds of self-training.
    \label{fig:ablation_smaples}}
    \vspace{-14pt}
\end{figure}
\mypara{Effect of different amount of unlabeled data.}
Customary to any unsupervised learning algorithm, we investigate how different amount of unlabeled data affects our algorithm. We randomly subsample our training set and report the performance of \ourmethod in \autoref{fig:ablation_smaples}. 
Across all ranges, we observe a general upward trend with more unlabeled data available. 
These results suggest that \ourmethod only needs a small amount of data to identify the rough location of mobile objects (evaluation at lower IoU is more lenient towards localization errors) and can significantly improve localization with more data.

\begin{table}[!t]
\centering
\renewcommand{\arraystretch}{1.05}
\small
\tabcolsep 2pt
\caption{\textbf{The precision and recall of the ``labels'' on the \lyft dataset ``train'' split.} 
We report the \textbf{\emph{precision\,/\,recall}} rate with BEV IoU=$0.25$ for mobile objects under various ranges. Please refer to \autoref{tbl:main} for naming.
\label{tbl:label_quality}}
\vspace{-8pt}
\resizebox{.47\textwidth}{!}{%
\begin{tabular}{=l|+C{37pt}|+C{37pt}|+C{37pt}||+C{37pt}}
\multicolumn{1}{c|}{\multirow{2}{*}{Method}} & \multicolumn{4}{c}{Precision\,/\,Recall @ IoU = 0.25} 
\\\cline{2-5}
 & 0-30 & 30-50 & 50-80 & 0-80 
 \\\hline
 \ourmethod-PP (seed)  
 & 56.3\,/\,63.8 & 21.7\,/\,41.1 & \phantom{0}8.6\,/\,\phantom{0}9.8 & 27.8\,/\,38.6 
 \\
 \ourmethod (seed)
 & 73.9\,/\,57.7 & 55.8\,/\,37.3 & 41.3\,/\,11.5 & 62.7\,/\,35.7 
 \\
 \ourmethod (R0)
 & 91.5\,/\,64.5 & 77.0\,/\,51.0 & 55.1\,/\,17.6 & 80.2\,/\,44.7 
 \\
 \ourmethod (R10)
 & 92.4\,/\,71.2 & 83.7\,/\,69.1 & 58.4\,/\,42.8 & 79.5\,/\,61.7 
 \\
  \ourmethod (R40)
 & 91.4\,/\,72.9 & 84.8\,/\,73.2 & 81.4\,/\,65.2 & 86.7\,/\,71.1
 \\
 \hline
\end{tabular}%
}
\vspace{-5pt}
\end{table}

\mypara{Quality of training labels.} 
The quality of the detector is determined by the quality of the automatically generated training labels.
We evaluate the generated pseudo-ground truth boxes in \autoref{tbl:label_quality} by computing their \emph{precision} and \emph{recall} compared to the ground truth.
For seed labels, compared with \ourmethod-PP, \pp score yields a set of labels with much higher precision but lower recall due to the filtering process.
This is in line with our intuition that these seed boxes are conservative but high quality.
After one round of training, the generated boxes have higher recall and precision.
Subsequent rounds of self-training substantially improves recall especially on the far range, affirming our intuition that the neural network slowly identifies missed objects that are consistent with the conservative training sets.
Put together, the whole process boosts the overall precision of seed labels by almost 40\% and nearly doubles overall recall.
This improved training data is reflected in the improvement in detector performance with self-training.
\begin{figure}
    \centering
    \begin{subfigure}{0.48\linewidth}
    \centering
    \includegraphics[width=\linewidth]{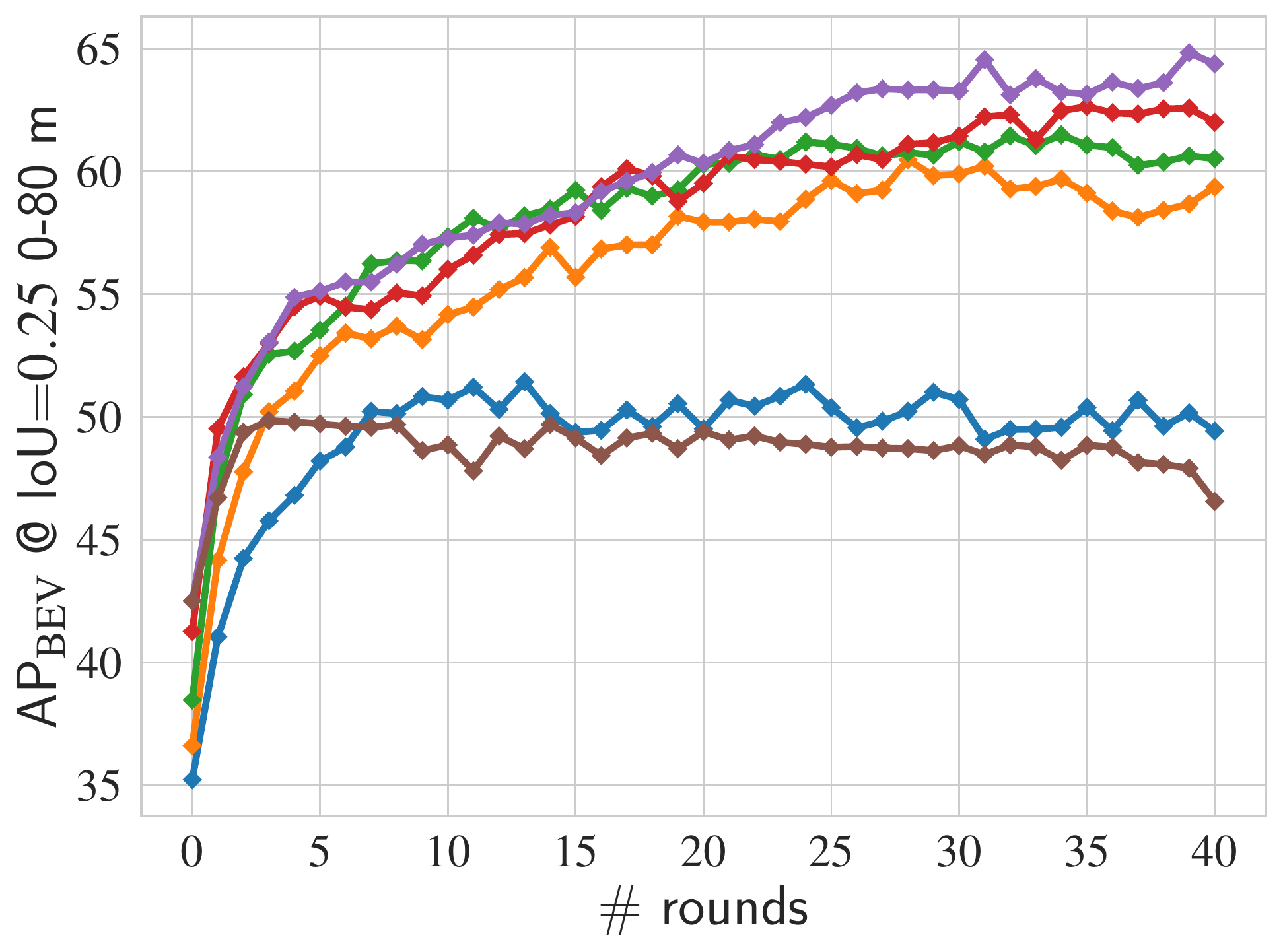}
    \caption{IoU=0.25 \label{fig:ablation_rounds_bev025}}
    \end{subfigure}
    \begin{subfigure}{0.48\linewidth}
    \centering
    \includegraphics[width=\linewidth]{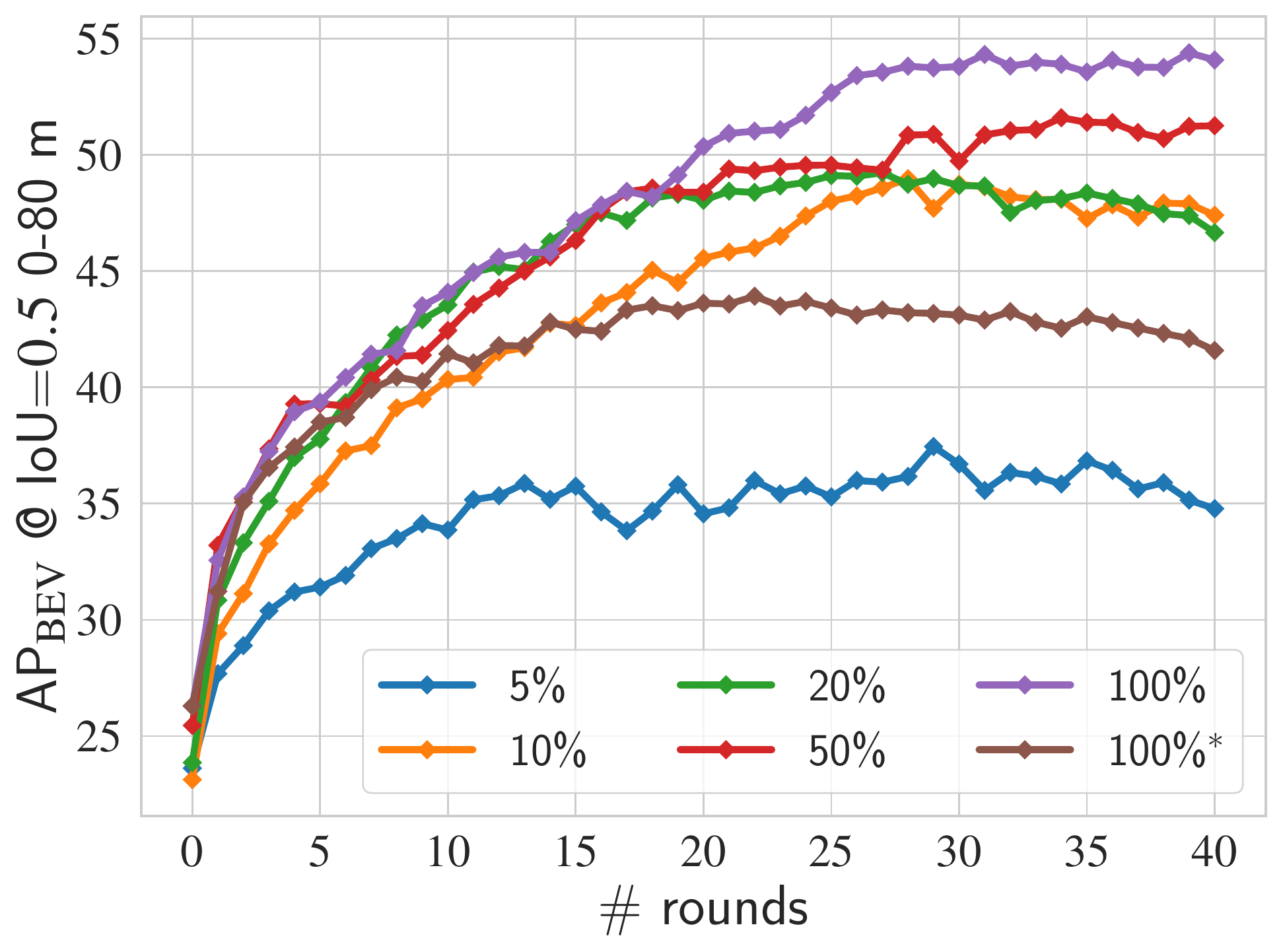}
    \caption{IoU=0.5 \label{fig:ablation_rounds_bev05}}
    \end{subfigure}
    \vspace{-8pt}
    \caption{\textbf{Number of self-training rounds \vs the performance.} We report \APBEV with IoU=0.5 and IoU=0.25 for mobile objects in 0-80\,m on \lyft test set from models trained with different rounds of self-training. We report results from models trained with different amount of unlabeled data. Note that the $100\%^*$ line is self-training without \pp score filtering. 
    \label{fig:ablation_rounds}
    }
    \vspace{-14pt}
\end{figure}

\mypara{Effect of different rounds of self-training.}
Given that self-training substantially improves the quality of the training labels, we next look at how this impacts the detector.
In \autoref{fig:ablation_rounds}, we show how \APBEV changes with different rounds of self-training. 
We observe that performance can improve for up to \textit{40 rounds} of self-training with larger amount of data ($\geq 50\%$).
Although more rounds can improve the performance, we emphasize that the improvement brought by \pp score filtering cannot be compensated by additional rounds of self-training (the $100\%^*$ line).

\begin{table}[!t]
\centering
\renewcommand{\arraystretch}{1.05}
\small
\tabcolsep 2pt
\caption{\textbf{Max recall with different methods on the \lyft dataset.} We report the max recall rate with BEV IoU$=0.25$ for mobile objects within 0-80m. Please refer to \autoref{tbl:main} for naming. Please refer to the supplementary for results on different depth ranges.
\label{tbl:recall_by_class_iou025_summary}}
\vspace{-8pt}
\resizebox{.4\textwidth}{!}{%
\begin{tabular}{=l|+C{24pt}|+C{24pt}|+C{24pt}|+C{24pt}}
\multicolumn{1}{c|}{Method} & Car & Truck & Ped. & Cyc. \\ \hline
\ourmethod-PP (R0) &  43.7 &  19.5 &  4.9 &  28.1\\
\ourmethod-PP (R10) &  52.4 &  35.1 & 0.5 &  13.2\\
\ourmethod (R0) & 54.6 &  31.8 & 10.2 & 50.5\\
\ourmethod (R10) & 72.6 &  46.8 & 42.5 & 60.7\\
\ourmethod (R40) & 81.4 &  46.4 & 42.1 & 62.9\\ \hline\hline
\rowstyle{\leavevmode\color{gray}}  Sup. (\kitti) &  73.6 & 49.4 & 33.5 & 56.7\\
\rowstyle{\leavevmode\color{gray}}  Sup. (\lyft)  &  83.6 & 65.4 & 53.8 & 67.5\\ \hline
\end{tabular}%
}
\vspace{-14pt}
\end{table}
\mypara{Maximum achievable recall by object types.}
In \autoref{tbl:recall_by_class_iou025_summary}, we further evaluate the maximum achievable recall of the different object types in the \lyft test set for various methods. 
We combine the \emph{other\_vehicle, truck, bus, emergency\_vehicle} in the raw \lyft dataset into the \emph{Truck} type and the \emph{motorcycle, bicycle} into the \emph{Cyclist} type. It can be seen that \ourmethod detects not only dominant, large objects in the dataset (Car), but also less common, smaller objects (Pedestrian and Cyclist).

\mypara{Common sense \vs self-training.} Clearly both our common sense-based seed boxes and our self-training approach are crucial for detector accuracy. But how do they stack against each other? In \autoref{tbl:ablation}, we attempt to trade-off the seed labels \vs the self-training by varying the number of scenes available to each step. We also experiment with switching off common sense-based (\ie, \pp score-based) filtering during self-training.
We observe that increasing the number of scenes for self-training has a bigger impact than increasing the size of the seed set (row 1 \vs rows 3 and 6).
Interestingly, without \pp score-based filtering, using 100\% data for seed labels performs worse than using only 5\% of the scenes. 
This may be because if all scenes are used for seed label computation and then used for training the detector, 
the detector may over-fit to the quirks of these labels and may not be able to correct them during self-training.
Having a ``held-out'' set of scenes for self-training thus seems beneficial.
We also observe that \pp score-based filtering does have a big impact on self-training and improves performance significantly (row 1 \vs 2, 4 \vs 5, 6 \vs 7).
Thus common sense-based filtering is crucial even within the self-training pipeline, suggesting a synergy between common sense and neural net training.

\mypara{Qualitative Results.} We show qualitative results of ``seed" label generation in \autoref{fig:seed_generation}, and visualization of self training on two scenes in \autoref{fig:teaser} on the ``train'' split of the \lyft dataset. Observe that the seed label generation filters out many of the superfluous clusters, but occasionally misses some objects or produces incorrectly sized objects. Via bootstrapping an object detector, our method can gradually recover the shape of mobile objects, as well as obtain higher recall than the initial seed label set (\autoref{fig:teaser}). Please refer to supplementary material for more.

\begin{table}[!t]
\centering
\renewcommand{\arraystretch}{1.05}
\small
\tabcolsep 2pt
\caption{\textbf{Common sense vs self-training.} 
We report \APBEV\,/\,\AP with IoU=$0.25$ for mobile objects under various ranges. Seed and ST column mean how much data are used as seed data and self-training data respectively; FT stands for filtering by \pp score during self-training. All are with 10 rounds of self-training.
\label{tbl:ablation}}
\vspace{-8pt}
\resizebox{.47\textwidth}{!}{%
\begin{tabular}{=C{21pt}|C{21pt}|C{10pt}|+C{38pt}|+C{38pt}|+C{38pt}||+C{38pt}} 
\multicolumn{3}{c|}{\multirow{1}{*}{Combinations}} & \multicolumn{4}{c}{\APBEV\,/\,\AP @ IoU = 0.25}\\
\hline
 Seed & ST & FT & 0-30 & 30-50 & 50-80 & 0-80 
 \\\hline
 5\% & 5\% & 
 & 54.8\,/\,53.2 & 40.4\,/\,39.4 & 17.3\,/\,16.3 & 39.4\,/\,37.6 
 \\
 5\% & 5\% & \checkmark 
 & 70.3\,/\,68.0 & 51.5\,/\,49.1 & 22.4\,/\,16.7 & 50.7\,/\,46.9 
 \\
 5\% & 100\% & 
 & 68.8\,/\,67.3 & 55.6\,/\,54.8 & 19.6\,/\,17.4 & 51.0\,/\,49.4 
 \\
 100\% & 5\% & 
 & 55.4\,/\,54.1 & 41.8\,/\,41.4 & 23.3\,/\,22.3 & 41.0\,/\,40.4 \\
 100\% & 5\% & \checkmark 
 & 68.0\,/\,65.6 & 49.0\,/\,47.4 & 28.7\,/\,25.3 & 51.1\,/\,47.9\\
 100\% & 100\% & 
 & 68.5\,/\,68.1 & 52.9\,/\,51.9 & 17.3\,/\,16.4 & 48.9\,/\,47.8 
 \\
 100\% & 100\% & \checkmark 
 & 73.8\,/\,71.3 & 62.8\,/\,60.3 & 27.0\,/\,24.8 & 57.3\,/\,55.1 
 \\
 
 \hline
\end{tabular}%
}
\vspace{-10pt}
\end{table}

\section{Discussion}
\label{sec:discussion}
\vspace{-0.5\baselineskip}

\mypara{Limitation.} Our approach focuses on learning to detect mobile objects and is evaluated by IoU between detections and ground-truth bounding boxes with single object type. We do not take the heading of objects into account, nor do we classify different object types. Also as mentioned above, our model tends to produce smaller boxes rather than amodal boxes. 
We leave these  as future work.

\mypara{Conclusion.} In this work, we explore a novel problem of learning a 3D mobile object detector from \lidar scans without any labels. Though this seems impossible at the first glance, we propose \ourmethod and show that with simple heuristics about mobile objects, \eg, they are not persistent over time, we can generate weak labels and bootstrap a surprisingly accurate detector from them. 
We evaluate \ourmethod exhaustively on two large-scale, real-world datasets and draw consistent conclusions.
We consider our work a first step towards a larger research effort to make entirely unsupervised object detectors a highly competitive reality. The potential impact of such an achievement would be monumental, allowing cars to be re-trained while in use, adapting to their local environments, enabling reliable driver assist and self-driving vehicles in developing countries, and avoiding privacy concerns by training vehicles on their own locally gathered data. We hope our work will inspire more research into this new and highly relevant problem.

\mypara{Acknowledgements}
This research is supported by grants from the National Science Foundation NSF (III-1618134, III-1526012, IIS-1149882, IIS-1724282, TRIPODS-1740822, IIS-2107077, OAC-2118240, OAC-2112606 and IIS-2107161), 
the Office of Naval Research DOD (N00014-17-1-2175), the DARPA Learning with Less Labels program (HR001118S0044), the Bill and Melinda Gates Foundation, the Cornell Center for Materials Research with funding from the NSF MRSEC program (DMR-1719875), and SAP America.
{\small
\bibliographystyle{ieee_fullname}
\bibliography{main.bib}

\begin{thebibliography}{10}\itemsep=-1pt

\bibitem{abbeloos20173d}
Wim Abbeloos, Esra Ataer-Cansizoglu, Sergio Caccamo, Yuichi Taguchi, and
  Yukiyasu Domae.
\newblock 3d object discovery and modeling using single rgb-d images containing
  multiple object instances.
\newblock In {\em 2017 International Conference on 3D Vision (3DV)}, pages
  431--439. IEEE, 2017.

\bibitem{arpit2017closer}
Devansh Arpit, Stanis{\l}aw Jastrz{\k{e}}bski, Nicolas Ballas, David Krueger,
  Emmanuel Bengio, Maxinder~S Kanwal, Tegan Maharaj, Asja Fischer, Aaron
  Courville, Yoshua Bengio, et~al.
\newblock A closer look at memorization in deep networks.
\newblock In {\em ICML}, 2017.

\bibitem{barnesephemerality}
Dan Barnes, Will Maddern, Geoffrey Pascoe, and Ingmar Posner.
\newblock Driven to distraction: Self-supervised distractor learning for robust
  monocular visual odometry in urban environments.
\newblock In {\em ICRA}, pages 1894--1900. IEEE, 2018.

\bibitem{berthelot2019remixmatch}
David Berthelot, Nicholas Carlini, Ekin~D Cubuk, Alex Kurakin, Kihyuk Sohn, Han
  Zhang, and Colin Raffel.
\newblock Remixmatch: Semi-supervised learning with distribution alignment and
  augmentation anchoring.
\newblock {\em Proceedings of the International Conference on Learning
  Representations}, 2020.

\bibitem{berthelot2019mixmatch}
David Berthelot, Nicholas Carlini, Ian Goodfellow, Nicolas Papernot, Avital
  Oliver, and Colin~A Raffel.
\newblock Mixmatch: A holistic approach to semi-supervised learning.
\newblock {\em Advances in Neural Information Processing Systems}, 32, 2019.

\bibitem{caesar2020nuscenes}
Holger Caesar, Varun Bankiti, Alex~H Lang, Sourabh Vora, Venice~Erin Liong,
  Qiang Xu, Anush Krishnan, Yu Pan, Giancarlo Baldan, and Oscar Beijbom.
\newblock nuscenes: A multimodal dataset for autonomous driving.
\newblock In {\em CVPR}, pages 11621--11631, 2020.

\bibitem{Argoverse}
Ming-Fang Chang, John~W Lambert, Patsorn Sangkloy, Jagjeet Singh, Slawomir Bak,
  Andrew Hartnett, De Wang, Peter Carr, Simon Lucey, Deva Ramanan, and James
  Hays.
\newblock Argoverse: 3d tracking and forecasting with rich maps.
\newblock In {\em Conference on Computer Vision and Pattern Recognition
  (CVPR)}, 2019.

\bibitem{chen2019progressive}
Chaoqi Chen, Weiping Xie, Wenbing Huang, Yu Rong, Xinghao Ding, Yue Huang,
  Tingyang Xu, and Junzhou Huang.
\newblock Progressive feature alignment for unsupervised domain adaptation.
\newblock In {\em Proceedings of the IEEE/CVF Conference on Computer Vision and
  Pattern Recognition}, pages 627--636, 2019.

\bibitem{chen20173d}
Xiaozhi Chen, Kaustav Kundu, Yukun Zhu, Huimin Ma, Sanja Fidler, and Raquel
  Urtasun.
\newblock 3d object proposals using stereo imagery for accurate object class
  detection.
\newblock {\em IEEE transactions on pattern analysis and machine intelligence},
  40(5):1259--1272, 2017.

\bibitem{chen2017multi}
Xiaozhi Chen, Huimin Ma, Ji Wan, Bo Li, and Tian Xia.
\newblock Multi-view {3D} object detection network for autonomous driving.
\newblock In {\em CVPR}, 2017.

\bibitem{cho2015}
Minsu Cho, Suha Kwak, Cordelia Schmid, and Jean Ponce.
\newblock Unsupervised object discovery and localization in the wild:
  Part-based matching with bottom-up region proposals.
\newblock In {\em Proceedings of the IEEE Conference on Computer Vision and
  Pattern Recognition}, 2015.

\bibitem{chong2013synthetic}
Zhuang~Jie Chong, Baoxing Qin, Tirthankar Bandyopadhyay, Marcelo~H Ang, Emilio
  Frazzoli, and Daniela Rus.
\newblock Synthetic 2d lidar for precise vehicle localization in 3d urban
  environment.
\newblock In {\em ICRA}, pages 1554--1559. IEEE, 2013.

\bibitem{choudhary2014slam}
Siddharth Choudhary, Alexander~JB Trevor, Henrik~I Christensen, and Frank
  Dellaert.
\newblock Slam with object discovery, modeling and mapping.
\newblock In {\em 2014 IEEE/RSJ International Conference on Intelligent Robots
  and Systems}, pages 1018--1025. IEEE, 2014.

\bibitem{collet2011structure}
Alvaro Collet, Siddhartha~S Srinivasay, and Martial Hebert.
\newblock Structure discovery in multi-modal data: a region-based approach.
\newblock In {\em 2011 IEEE International Conference on Robotics and
  Automation}, pages 5695--5702. IEEE, 2011.

\bibitem{collet2013exploiting}
Alvaro Collet, Bo Xiong, Corina Gurau, Martial Hebert, and Siddhartha~S
  Srinivasa.
\newblock Exploiting domain knowledge for object discovery.
\newblock In {\em 2013 IEEE International Conference on Robotics and
  Automation}, pages 2118--2125. IEEE, 2013.

\bibitem{collet2015herbdisc}
Alvaro Collet, Bo Xiong, Corina Gurau, Martial Hebert, and Siddhartha~S
  Srinivasa.
\newblock Herbdisc: Towards lifelong robotic object discovery.
\newblock {\em The International Journal of Robotics Research}, 34(1):3--25,
  2015.

\bibitem{du2020unsupervised}
Yilun Du, Kevin Smith, Tomer Ulman, Joshua Tenenbaum, and Jiajun Wu.
\newblock Unsupervised discovery of 3d physical objects from video.
\newblock {\em arXiv preprint arXiv:2007.12348}, 2020.

\bibitem{ester1996density}
Martin Ester, Hans-Peter Kriegel, J{\"o}rg Sander, Xiaowei Xu, et~al.
\newblock A density-based algorithm for discovering clusters in large spatial
  databases with noise.
\newblock In {\em kdd}, volume~96, pages 226--231, 1996.

\bibitem{faktor2013co}
Alon Faktor and Michal Irani.
\newblock Co-segmentation by composition.
\newblock In {\em Proceedings of the IEEE international conference on computer
  vision}, pages 1297--1304, 2013.

\bibitem{garcia2015saliency}
Germ{\'a}n~M Garc{\'\i}a, Ekaterina Potapova, Thomas Werner, Michael Zillich,
  Markus Vincze, and Simone Frintrop.
\newblock Saliency-based object discovery on rgb-d data with a late-fusion
  approach.
\newblock In {\em 2015 IEEE International Conference on Robotics and Automation
  (ICRA)}, pages 1866--1873. IEEE, 2015.

\bibitem{geiger2013vision}
Andreas Geiger, Philip Lenz, Christoph Stiller, and Raquel Urtasun.
\newblock Vision meets robotics: The kitti dataset.
\newblock {\em The International Journal of Robotics Research},
  32(11):1231--1237, 2013.

\bibitem{geiger2012we}
Andreas Geiger, Philip Lenz, and Raquel Urtasun.
\newblock Are we ready for autonomous driving? the kitti vision benchmark
  suite.
\newblock In {\em CVPR}, 2012.

\bibitem{Ghiasi_2021_multi_ST}
Golnaz Ghiasi, Barret Zoph, Ekin~D. Cubuk, Quoc~V. Le, and Tsung-Yi Lin.
\newblock Multi-task self-training for learning general representations.
\newblock In {\em Proceedings of the IEEE/CVF International Conference on
  Computer Vision (ICCV)}, pages 8856--8865, October 2021.

\bibitem{grandvalet2005entropy}
Yves Grandvalet, Yoshua Bengio, et~al.
\newblock Semi-supervised learning by entropy minimization.
\newblock {\em CAP}, 367:281--296, 2005.

\bibitem{han2018co}
Bo Han, Quanming Yao, Xingrui Yu, Gang Niu, Miao Xu, Weihua Hu, Ivor Tsang, and
  Masashi Sugiyama.
\newblock Co-teaching: Robust training of deep neural networks with extremely
  noisy labels.
\newblock In {\em NeurIPS}, 2018.

\bibitem{herbst2011toward}
Evan Herbst, Peter Henry, Xiaofeng Ren, and Dieter Fox.
\newblock Toward object discovery and modeling via 3-d scene comparison.
\newblock In {\em 2011 IEEE International Conference on Robotics and
  Automation}, pages 2623--2629. IEEE, 2011.

\bibitem{herbst2011rgb}
Evan Herbst, Xiaofeng Ren, and Dieter Fox.
\newblock Rgb-d object discovery via multi-scene analysis.
\newblock In {\em 2011 IEEE/RSJ International Conference on Intelligent Robots
  and Systems}, pages 4850--4856. IEEE, 2011.

\bibitem{karpathy2013object}
Andrej Karpathy, Stephen Miller, and Li Fei-Fei.
\newblock Object discovery in 3d scenes via shape analysis.
\newblock In {\em 2013 IEEE International Conference on Robotics and
  Automation}, pages 2088--2095. IEEE, 2013.

\bibitem{lyft2019}
R. Kesten, M. Usman, J. Houston, T. Pandya, K. Nadhamuni, A. Ferreira, M. Yuan,
  B. Low, A. Jain, P. Ondruska, S. Omari, S. Shah, A. Kulkarni, A. Kazakova, C.
  Tao, L. Platinsky, W. Jiang, and V. Shet.
\newblock Lyft level 5 av dataset 2019.
\newblock \url{https://level5.lyft.com/dataset/}, 2019.

\bibitem{kochanov2016scene}
Deyvid Kochanov, Aljo{\v{s}}a O{\v{s}}ep, J{\"o}rg St{\"u}ckler, and Bastian
  Leibe.
\newblock Scene flow propagation for semantic mapping and object discovery in
  dynamic street scenes.
\newblock In {\em 2016 IEEE/RSJ International Conference on Intelligent Robots
  and Systems (IROS)}, pages 1785--1792. IEEE, 2016.

\bibitem{lang2019pointpillars}
Alex~H Lang, Sourabh Vora, Holger Caesar, Lubing Zhou, Jiong Yang, and Oscar
  Beijbom.
\newblock Pointpillars: Fast encoders for object detection clouds.
\newblock In {\em ICCV}, pages 12697--12705, 2019.

\bibitem{lee2013pseudo}
Dong-Hyun Lee et~al.
\newblock Pseudo-label: The simple and efficient semi-supervised learning
  method for deep neural networks.
\newblock In {\em Workshop on challenges in representation learning, ICML},
  volume~3, page 896, 2013.

\bibitem{li2018deep}
Weihao Li, Omid~Hosseini Jafari, and Carsten Rother.
\newblock Deep object co-segmentation.
\newblock In {\em Asian Conference on Computer Vision}, pages 638--653.
  Springer, 2018.

\bibitem{liang2020learning}
Ming Liang, Bin Yang, Rui Hu, Yun Chen, Renjie Liao, Song Feng, and Raquel
  Urtasun.
\newblock Learning lane graph representations for motion forecasting, 2020.

\bibitem{ma2015simultaneous}
Lu Ma, Mahsa Ghafarianzadeh, David Coleman, Nikolaus Correll, and Gabe Sibley.
\newblock Simultaneous localization, mapping, and manipulation for unsupervised
  object discovery.
\newblock In {\em 2015 IEEE International Conference on Robotics and Automation
  (ICRA)}, pages 1344--1351. IEEE, 2015.

\bibitem{ma2014unsupervised}
Lu Ma and Gabe Sibley.
\newblock Unsupervised dense object discovery, detection, tracking and
  reconstruction.
\newblock In {\em European Conference on Computer Vision}, pages 80--95.
  Springer, 2014.

\bibitem{mason2012object}
Julian Mason, Bhaskara Marthi, and Ronald Parr.
\newblock Object disappearance for object discovery.
\newblock In {\em 2012 IEEE/RSJ International Conference on Intelligent Robots
  and Systems}, pages 2836--2843. IEEE, 2012.

\bibitem{mason2014unsupervised}
Julian Mason, Bhaskara Marthi, and Ronald Parr.
\newblock Unsupervised discovery of object classes with a mobile robot.
\newblock In {\em 2014 IEEE International Conference on Robotics and Automation
  (ICRA)}, pages 3074--3081. IEEE, 2014.

\bibitem{mei2020instance}
Ke Mei, Chuang Zhu, Jiaqi Zou, and Shanghang Zhang.
\newblock Instance adaptive self-training for unsupervised domain adaptation.
\newblock In {\em Proceedings of the European Conference on Computer Vision
  (ECCV)}, 2020.

\bibitem{pan2021model}
Tai-Yu Pan, Cheng Zhang, Yandong Li, Hexiang Hu, Dong Xuan, Soravit Changpinyo,
  Boqing Gong, and Wei-Lun Chao.
\newblock On model calibration for long-tailed object detection and instance
  segmentation.
\newblock In {\em NeurIPS}, 2021.

\bibitem{Pathak2017Learning}
Deepak Pathak, Ross Girshick, Piotr Doll\'{a}r, Trevor Darrell, and Bharath
  Hariharan.
\newblock Learning features by watching objects move.
\newblock In {\em CVPR}, 2017.

\bibitem{Phoo_2021_ICCV}
Cheng~Perng Phoo and Bharath Hariharan.
\newblock Coarsely-labeled data for better few-shot transfer.
\newblock In {\em Proceedings of the IEEE/CVF International Conference on
  Computer Vision (ICCV)}, pages 9052--9061, October 2021.

\bibitem{phoo2021STARTUP}
Cheng~Perng Phoo and Bharath Hariharan.
\newblock Self-training for few-shot transfer across extreme task differences.
\newblock In {\em Proceedings of the International Conference on Learning
  Representations}, 2021.

\bibitem{pleiss2020identifying}
Geoff Pleiss, Tianyi Zhang, Ethan~R Elenberg, and Kilian~Q Weinberger.
\newblock Identifying mislabeled data using the area under the margin ranking.
\newblock In {\em NeurIPS}, 2020.

\bibitem{qi2018frustum}
Charles~R Qi, Wei Liu, Chenxia Wu, Hao Su, and Leonidas~J Guibas.
\newblock Frustum pointnets for 3d object detection from rgb-d data.
\newblock In {\em CVPR}, 2018.

\bibitem{qi2017pointnet}
Charles~R Qi, Hao Su, Kaichun Mo, and Leonidas~J Guibas.
\newblock Pointnet: Deep learning on point sets for 3d classification and
  segmentation.
\newblock In {\em CVPR}, 2017.

\bibitem{qi2017pointnet++}
Charles~Ruizhongtai Qi, Li Yi, Hao Su, and Leonidas~J Guibas.
\newblock Pointnet++: Deep hierarchical feature learning on point sets in a
  metric space.
\newblock In {\em NeurIPS}, 2017.

\bibitem{shi2020pv}
Shaoshuai Shi, Chaoxu Guo, Li Jiang, Zhe Wang, Jianping Shi, Xiaogang Wang, and
  Hongsheng Li.
\newblock Pv-rcnn: Point-voxel feature set abstraction for 3d object detection.
\newblock In {\em CVPR}, pages 10529--10538, 2020.

\bibitem{shi2019pointrcnn}
Shaoshuai Shi, Xiaogang Wang, and Hongsheng Li.
\newblock Pointrcnn: 3d object proposal generation and detection from point
  cloud.
\newblock In {\em CVPR}, 2019.

\bibitem{shin2010unsupervised}
Jiwon Shin, Rudolph Triebel, and Roland Siegwart.
\newblock Unsupervised discovery of repetitive objects.
\newblock In {\em 2010 IEEE International Conference on Robotics and
  Automation}, pages 5041--5046. IEEE, 2010.

\bibitem{sohn2020fixmatch}
Kihyuk Sohn, David Berthelot, Chun-Liang Li, Zizhao Zhang, Nicholas Carlini,
  Ekin~D Cubuk, Alex Kurakin, Han Zhang, and Colin Raffel.
\newblock Fixmatch: Simplifying semi-supervised learning with consistency and
  confidence.
\newblock {\em Advances in Neural Information Processing Systems}, 2020.

\bibitem{openpcdet2020}
OpenPCDet~Development Team.
\newblock Openpcdet: An open-source toolbox for 3d object detection clouds.
\newblock \url{https://github.com/open-mmlab/OpenPCDet}, 2020.

\bibitem{tian2021unsupervised}
Hao Tian, Yuntao Chen, Jifeng Dai, Zhaoxiang Zhang, and Xizhou Zhu.
\newblock Unsupervised object detection with lidar clues.
\newblock In {\em Proceedings of the IEEE/CVF Conference on Computer Vision and
  Pattern Recognition}, pages 5962--5972, 2021.

\bibitem{tian2020rethinking}
Yonglong Tian, Yue Wang, Dilip Krishnan, Joshua~B Tenenbaum, and Phillip Isola.
\newblock Rethinking few-shot image classification: a good embedding is all you
  need?
\newblock In {\em Computer Vision--ECCV 2020: 16th European Conference,
  Glasgow, UK, August 23--28, 2020, Proceedings, Part XIV 16}, pages 266--282.
  Springer, 2020.

\bibitem{triebel2010segmentation}
Rudolph Triebel, Jiwon Shin, and Roland Siegwart.
\newblock Segmentation and unsupervised part-based discovery of repetitive
  objects.
\newblock {\em Robotics: Science and Systems VI}, pages 1--8, 2010.

\bibitem{vo2019unsupervised}
Huy~V Vo, Francis Bach, Minsu Cho, Kai Han, Yann LeCun, Patrick P{\'e}rez, and
  Jean Ponce.
\newblock Unsupervised image matching and object discovery as optimization.
\newblock In {\em Proceedings of the IEEE/CVF Conference on Computer Vision and
  Pattern Recognition}, pages 8287--8296, 2019.

\bibitem{vo2020toward}
Huy~V Vo, Patrick P{\'e}rez, and Jean Ponce.
\newblock Toward unsupervised, multi-object discovery in large-scale image
  collections.
\newblock In {\em European Conference on Computer Vision}, pages 779--795.
  Springer, 2020.

\bibitem{wang2020train}
Yan Wang, Xiangyu Chen, Yurong You, Li~Erran Li, Bharath Hariharan, Mark
  Campbell, Kilian~Q. Weinberger, and Wei-Lun Chao.
\newblock Train in germany, test in the usa: Making 3d object detectors
  generalize.
\newblock In {\em CVPR}, pages 11713--11723, June 2020.

\bibitem{xie2020self}
Qizhe Xie, Minh-Thang Luong, Eduard Hovy, and Quoc~V Le.
\newblock Self-training with noisy student improves imagenet classification.
\newblock In {\em Proceedings of the IEEE/CVF Conference on Computer Vision and
  Pattern Recognition}, pages 10687--10698, 2020.

\bibitem{yan2018second}
Yan Yan, Yuxing Mao, and Bo Li.
\newblock Second: Sparsely embedded convolutional detection.
\newblock {\em Sensors}, 18(10):3337, 2018.

\bibitem{yang2018pixor}
Bin Yang, Wenjie Luo, and Raquel Urtasun.
\newblock Pixor: Real-time 3d object detection from point clouds.
\newblock In {\em CVPR}, 2018.

\bibitem{yang2021st3d}
Jihan Yang, Shaoshuai Shi, Zhe Wang, Hongsheng Li, and Xiaojuan Qi.
\newblock St3d: Self-training for unsupervised domain adaptation on 3d object
  detection.
\newblock In {\em Proceedings of the IEEE/CVF Conference on Computer Vision and
  Pattern Recognition}, 2021.

\bibitem{yang20203dssd}
Zetong Yang, Yanan Sun, Shu Liu, and Jiaya Jia.
\newblock 3dssd: Point-based 3d single stage object detector.
\newblock In {\em CVPR}, pages 11040--11048, 2020.

\bibitem{yuan2017deep}
Ze-Huan Yuan, Tong Lu, Yirui Wu, et~al.
\newblock Deep-dense conditional random fields for object co-segmentation.
\newblock In {\em IJCAI}, pages 3371--3377, 2017.

\bibitem{zhang2021understanding}
Chiyuan Zhang, Samy Bengio, Moritz Hardt, Benjamin Recht, and Oriol Vinyals.
\newblock Understanding deep learning (still) requires rethinking
  generalization.
\newblock {\em Communications of the ACM}, 64(3):107--115, 2021.

\bibitem{zhang2013unsupervised}
Quanshi Zhang, Xuan Song, Xiaowei Shao, Huijing Zhao, and Ryosuke Shibasaki.
\newblock Unsupervised 3d category discovery and point labeling from a large
  urban environment.
\newblock In {\em 2013 IEEE International Conference on Robotics and
  Automation}, pages 2685--2692. IEEE, 2013.

\bibitem{zhang2019category}
Qiming Zhang, Jing Zhang, Wei Liu, and Dacheng Tao.
\newblock Category anchor-guided unsupervised domain adaptation for semantic
  segmentation.
\newblock In {\em Advances in neural information processing systems}, 2019.

\bibitem{zhang2017efficient}
Xiao Zhang, Wenda Xu, Chiyu Dong, and John~M Dolan.
\newblock Efficient l-shape fitting for vehicle detection using laser scanners.
\newblock In {\em 2017 IEEE Intelligent Vehicles Symposium (IV)}, pages 54--59.
  IEEE, 2017.

\bibitem{zhou2019end}
Yin Zhou, Pei Sun, Yu Zhang, Dragomir Anguelov, Jiyang Gao, Tom Ouyang, James
  Guo, Jiquan Ngiam, and Vijay Vasudevan.
\newblock End-to-end multi-view fusion for {3D} object detection in lidar point
  clouds.
\newblock In {\em CoRL}, 2020.

\bibitem{zhou2018voxelnet}
Yin Zhou and Oncel Tuzel.
\newblock Voxelnet: End-to-end learning for point cloud based 3d object
  detection.
\newblock In {\em CVPR}, 2018.

\bibitem{zou2018unsupervised}
Yang Zou, Zhiding Yu, BVK Kumar, and Jinsong Wang.
\newblock Unsupervised domain adaptation for semantic segmentation via
  class-balanced self-training.
\newblock In {\em Proceedings of the European conference on computer vision
  (ECCV)}, pages 289--305, 2018.

\bibitem{zou2019confidence}
Yang Zou, Zhiding Yu, Xiaofeng Liu, BVK Kumar, and Jinsong Wang.
\newblock Confidence regularized self-training.
\newblock In {\em Proceedings of the IEEE/CVF International Conference on
  Computer Vision}, pages 5982--5991, 2019.

\end{thebibliography}
}

\newpage
\setcounter{equation}{0}
\setcounter{figure}{0}
\setcounter{table}{0}
\setcounter{page}{1}
\setcounter{section}{0}
\renewcommand{\thesection}{S\arabic{section}}
\renewcommand{\thetable}{S\arabic{table}}
\renewcommand{\thefigure}{S\arabic{figure}}

\begin{center}
  \textbf{\Large Supplementary Material}
\end{center}

\section{Implementation details}
We set $[-H_s, H_e]$ to $[0, 70]$\,m since we experiment with frontal-view detection only. We combine only one scan into the dense point cloud $\cptc_c^t$ every 2~m within this range. In calculating \pp score, we use as many traversals as possible ($\geq 2$) and set $r=0.3$m. For clustering, we use $K=70$ and $r'=2.0$m in the graph, and $\epsilon=0.1$, min\_samples~$=10$ for DBSCAN. For filtering, we use a loose threshold of $\alpha=20$ percentile and $\gamma=0.7$. Other common sense properties are simply implemented as follows:
\begin{itemize}[itemsep=1pt,topsep=3pt]
    \item $\#$ points in the cluster $>= 10$;
    \item Volume of fitted bounding boxes $\in [0.5, 120]m^3$;
    \item The height (upright distance against the ground plane) of points $\text{Height}_{\max} > 0.5m$ and $\text{Height}_{\min} < 1.0m$ to ensure clusters not floating in the air or beneath the ground due to errors in \lidar.
\end{itemize}
We did not tune these parameters except qualitatively checked the fitted bounding boxes in few scenes in the \lyft ``train'' set. 
For detection models, we use the default hyper-parameters tuned on KITTI\footnote{\url{https://github.com/open-mmlab/OpenPCDet/tree/master/tools/cfgs/kitti_models}} with few exceptions listed in the paper. We will open-source the code upon acceptance.

\section{Experiments with other detectors}
Besides the PointRCNN detector~\cite{shi2019pointrcnn}, We experiment with two other detectors PointPillars~\cite{lang2019pointpillars} and VoxelNet (SECOND)~\cite{zhou2018voxelnet, yan2018second}, and show their results in \autoref{tbl:second} and \autoref{tbl:pointpillars}. We apply the default hyper-parameters of these two models tuned on KITTI, and apply the same procedure as that on PointRCNN models. Note that PointPillars and VoxelNet model need a pre-defined anchor size for different types of objects, which we picked (length, width, height) as ($2.0, 1.0, 1.7$)\,m without tuning. We observe that generally the PointPillars and VoxelNet yield worse results than PointRCNN models (possibly due to the fixed anchor size for all mobile objects), but we still observe significant gains from self-training.

\begin{table*}[!t]
\centering
\renewcommand{\arraystretch}{1.05}
\tabcolsep 2pt
\caption{\textbf{Detection performance with PointPillars~\cite{lang2019pointpillars} on the \lyft dataset.} 
Please refer to \autoref{tbl:main} for naming.
\label{tbl:pointpillars}}
\begin{tabular}{=l|+C{44pt}|+C{44pt}|+C{44pt}||+C{44pt}||+C{44pt}|+C{44pt}|+C{44pt}||+C{44pt}}
\multicolumn{1}{c|}{\multirow{2}{*}{Method}} & \multicolumn{4}{c||}{\APBEV / \AP @ IoU = 0.25} &  \multicolumn{4}{c}{\APBEV / \AP @ IoU = 0.5}\\
\cline{2-9}\
 & 0-30 & 30-50 & 50-80 & 0-80 & 0-30 & 30-50 & 50-80 & 0-80\\\hline
 \ourmethod (R0)
 & 56.3 / 51.3 & 26.6 / 19.5 & \phantom{0}5.4 / \phantom{0}3.0 & 30.4 / 24.6  
 & 32.1 / 25.2 & 10.0 / \phantom{0}4.2 & \phantom{0}1.2 / \phantom{0}0.2 & 13.8 / \phantom{0}8.5 \\
 \ourmethod (R10)
 & 55.7 / 49.1 & 43.1 / 38.4 & \phantom{0}8.8 / \phantom{0}7.5 & 33.9 / 29.9
 & 37.4 / 27.7 & 28.8 / 10.0 & \phantom{0}5.0 / \phantom{0}1.1 & 22.1 / 10.7\\
\hline\hline
\rowstyle{\leavevmode\color{gray}} Sup. (\lyft)
& 78.7 / 77.9 & 64.6 / 63.7 & 45.4 / 44.1 & 64.7 / 63.6 
& 72.9 / 68.9 & 55.5 / 50.3 & 41.5 / 35.4 & 58.0 / 52.8 \\
 \hline
\end{tabular}
\end{table*}

\begin{table*}[!t]
\centering
\renewcommand{\arraystretch}{1.05}
\tabcolsep 2pt
\caption{\textbf{Detection performance with VoxelNet (SECOND)~\cite{zhou2018voxelnet, yan2018second} on the \lyft dataset.} 
Please refer to \autoref{tbl:main} for naming.
\label{tbl:second}}
\begin{tabular}{=l|+C{44pt}|+C{44pt}|+C{44pt}||+C{44pt}||+C{44pt}|+C{44pt}|+C{44pt}||+C{44pt}}
\multicolumn{1}{c|}{\multirow{2}{*}{Method}} & \multicolumn{4}{c||}{\APBEV / \AP @ IoU = 0.25} &  \multicolumn{4}{c}{\APBEV / \AP @ IoU = 0.5}\\
\cline{2-9}\
 & 0-30 & 30-50 & 50-80 & 0-80 & 0-30 & 30-50 & 50-80 & 0-80\\\hline
 \ourmethod (R0)
 & 54.3 / 49.7 & 27.8 / 21.4 & \phantom{0}4.9 / \phantom{0}2.8 & 30.2 / 24.8
 & 30.3 / 24.9 & 11.7 / \phantom{0}5.1 & \phantom{0}1.1 / \phantom{0}0.2 & 14.0 / \phantom{0}8.8 \\
 \ourmethod (R10)
 & 54.9 / 44.8 & 38.7 / 31.5 & \phantom{0}8.3 / \phantom{0}6.2 & 32.5 / 26.0
 & 32.1 / 24.3 & 20.0 / \phantom{0}7.7 & \phantom{0}3.4 / \phantom{0}0.8 & 17.0 / \phantom{0}8.9\\
\hline\hline
\rowstyle{\leavevmode\color{gray}} Sup. (\lyft)
& 81.6 / 81.1 & 67.8 / 66.3 & 45.5 / 44.6 & 65.9 / 64.9
& 76.7 / 73.9 & 59.7 / 55.3 & 41.8 / 36.3 & 60.1 / 55.4 \\
 \hline
\end{tabular}
\end{table*}

\begin{table*}[!t]
\centering
\renewcommand{\arraystretch}{1.05}
\tabcolsep 2pt
\caption{\textbf{Max recall with different methods on the \lyft dataset.} 
Please refer to \autoref{tbl:main} for naming.
This corresponds to the counterpart \autoref{tbl:recall_by_class_iou025_summary} in the main paper.
\label{tbl:recall_by_class}}

\begin{subtable}{\textwidth}

\caption{Recall @ IoU=0.5}
\resizebox{.99\textwidth}{!}{%
\begin{tabular}{=l|+C{24pt}|+C{24pt}|+C{24pt}|+C{24pt}||+C{24pt}|+C{24pt}|+C{24pt}|+C{24pt}||+C{24pt}|+C{24pt}|+C{24pt}|+C{24pt}||+C{24pt}|+C{24pt}|+C{24pt}|+C{24pt}}
\multicolumn{1}{c|}{\multirow{2}{*}{Method}} & \multicolumn{4}{c||}{Car} & \multicolumn{4}{c||}{Truck} & \multicolumn{4}{c||}{Pedestrian} & \multicolumn{4}{c}{Cyclist} \\ \cline{2-17} 
 & 0-30 & 30-50 & 50-80 & 0-80 & 0-30 & 30-50 & 50-80 & 0-80 & 0-30 & 30-50 & 50-80 & 0-80 & 0-30 & 30-50 & 50-80 & 0-80 \\ \hline
\ourmethod-PP (R0) & 57.6 & 27.3  & 3.0   & 30.1 & 36.1 & 5.0   & 0.2   & 9.1  & 0.9  & 1.1   & 0.8   & 1.0  & 14.4 & 10.7  & 2.1   & 10.7 \\
\ourmethod-PP (R10) & 63.2 & 49.1  & 8.0   & 40.9 & 39.0 & 21.4  & 5.6   & 20.1 & 0.0  & 0.0   & 0.2   & 0.1  & 8.1  & 11.5  & 2.9   & 8.2  \\
\ourmethod (R0) & 63.7 & 45.5  & 11.2  & 41.0 & 29.8 & 18.3  & 2.4   & 17.2 & 5.2  & 0.6   & 0.0   & 1.7  & 34.1 & 11.7  & 1.2   & 20.2 \\
\ourmethod (R10) & 67.5 & 70.7  & 40.9  & 60.6 & 35.1 & 28.2  & 13.3  & 25.3 & 35.1 & 35.0  & 7.6   & 27.5 & 62.9 & 41.1  & 7.1   & 44.7 \\
\ourmethod (R40) & 69.9 & 78.8 & 68.9 & 72.9 & 25.4 & 27.4 & 20.1 & 28.0 & 39.4 & 38.8 & 6.2 & 28.6 & 70.6 & 32.5 & 1.5 & 44.4 \\ \hline\hline
\rowstyle{\leavevmode\color{gray}}  Sup. (\kitti) & 82.1 & 76.3  & 53.3  & 71.2 & 45.9 & 23.8  & 23.3  & 30.0 & 56.9 & 32.1  & 2.5   & 29.7 & 59.5 & 16.1  & 1.2   & 33.8 \\
\rowstyle{\leavevmode\color{gray}}  Sup. (\lyft) & 85.7 & 82.5  & 75.2  & 81.5 & 64.9 & 52.0  & 50.3  & 54.7 & 60.8 & 55.4  & 18.9  & 45.2 & 71.0 & 43.2  & 6.8   & 49.2 \\ \hline
\end{tabular}%
}
\end{subtable}
\vspace{10pt}

\begin{subtable}{\textwidth}
\caption{Recall @ IoU=0.25}
\resizebox{.99\textwidth}{!}{%
\begin{tabular}{=l|+C{24pt}|+C{24pt}|+C{24pt}|+C{24pt}||+C{24pt}|+C{24pt}|+C{24pt}|+C{24pt}||+C{24pt}|+C{24pt}|+C{24pt}|+C{24pt}||+C{24pt}|+C{24pt}|+C{24pt}|+C{24pt}}
\multicolumn{1}{c|}{\multirow{2}{*}{Method}} & \multicolumn{4}{c||}{Car} & \multicolumn{4}{c||}{Truck} & \multicolumn{4}{c||}{Pedestrian} & \multicolumn{4}{c}{Cyclist} \\ \cline{2-17} 
  & 0-30 & 30-50 & 50-80 & 0-80 & 0-30 & 30-50 & 50-80 & 0-80 & 0-30 & 30-50 & 50-80 & 0-80 & 0-30 & 30-50 & 50-80 & 0-80 \\ \hline
\ourmethod-PP (R0) &  65.5 & 53.0 & 8.8 & 43.7 & 52.2 & 19.1 & 1.9 & 19.5 & 2.8 & 6.1 & 4.9 & 4.9 & 33.9 & 31.7 & 8.3 & 28.1\\
\ourmethod-PP (R10) &  73.0 & 66.2 & 14.0 & 52.4 & 62.9 & 36.0 & 9.2 & 35.1 & 0.9 & 0.5 & 0.2 & 0.5 & 15.5 & 15.3 & 4.1 & 13.2\\
\ourmethod (R0) &  73.9 & 63.6 & 22.9 & 54.6 & 46.8 & 30.5 & 12.2 & 31.8 & 22.5 & 8.6 & 1.5 & 10.2 & 75.3 & 42.3 & 4.4 & 50.5\\
\ourmethod (R10) &  79.0 & 80.8 & 55.1 & 72.6 & 59.5 & 43.1 & 33.6 & 46.8 & 61.7 & 51.4 & 14.6 & 42.5 & 71.8 & 66.0 & 26.0 & 60.7\\
\ourmethod (R40) &  81.2 & 83.1 & 78.1 & 81.4 & 53.2 & 43.3 & 31.7 & 46.4 & 63.9 & 51.8 & 11.0 & 42.1 & 81.6 & 62.0 & 19.8 & 62.9\\ \hline\hline
\rowstyle{\leavevmode\color{gray}}  Sup. (\kitti) & 82.8 & 78.3 & 57.5 & 73.6 & 71.7 & 42.3 & 33.4 & 49.4 & 66.0 & 35.0 & 2.8 & 33.5 & 84.1 & 45.3 & 9.7 & 56.7\\
\rowstyle{\leavevmode\color{gray}}  Sup. (\lyft)  &  86.9 & 84.1 & 78.5 & 83.6 & 73.7 & 66.3 & 58.0 & 65.4 & 72.3 & 63.7 & 25.2 & 53.8 & 83.1 & 68.1 & 29.2 & 67.5\\ \hline
\end{tabular}%
}
\end{subtable}
\vspace{-10pt}
\end{table*}

\begin{table}[!t]
\centering
\renewcommand{\arraystretch}{1.05}
\tabcolsep 2pt
\caption{\textbf{Detection performance with different methods on the \lyft dataset.} 
We report \APBEV / \AP with IoU=$0.5$ for mobile objects under various ranges.
Please refer to \autoref{tbl:main} for naming.
This corresponds to the counterpart \autoref{tbl:main} in the main paper.
\vspace{-8pt}
\label{tbl:main_iou05}}
\resizebox{.47\textwidth}{!}{%
\begin{tabular}{=l|+C{44pt}|+C{44pt}|+C{44pt}||+C{44pt}}
\multicolumn{1}{c|}{\multirow{2}{*}{Method}} & \multicolumn{4}{c}{\APBEV / \AP @ IoU = 0.25}\\
\cline{2-5}\
 & 0-30 & 30-50 & 50-80 & 0-80 \\\hline
 \ourmethod-\pp (R0)
 & 34.1 / 31.3 & \phantom{0}5.1 / \phantom{0}3.0 & \phantom{0}0.0 / \phantom{0}0.0 & 12.0 / \phantom{0}9.7 
\\
 \ourmethod-\pp (R10)
 & 42.1 / 38.3 & 21.9 / 19.2 & \phantom{0}1.0 / \phantom{0}0.9 & 22.8 / 20.6
 \\
 \ourmethod (R0)
 & 44.9 / 40.4 & 24.5 / 14.8 & \phantom{0}2.7 / \phantom{0}0.7 & 26.3 / 19.8 
 \\
 \ourmethod (R10)
 & 56.8 / 51.3 & 51.4 / 40.5 & 19.2 / \phantom{0}9.0 & 44.1 / 35.5 
 \\
 \ourmethod (R40)
 & 61.1 / 56.2 & 57.5 / 53.4 & 41.2 / 29.8 & 54.1 / 47.6 
 \\
\hline\hline
\rowstyle{\leavevmode\color{gray}} Sup. (\kitti) 
& 72.3 / 69.5 & 53.2 / 48.1 & 27.9 / 20.5 & 53.1 / 48.1  
\\
\rowstyle{\leavevmode\color{gray}} Sup. (\lyft)
& 79.6 / 77.5 & 66.4 / 64.4& 47.8 / 43.8 & 65.5 / 63.2  
\\
 \hline
\end{tabular}
}
\vspace{-8pt}
\end{table}

\begin{table}[!t]
\centering
\renewcommand{\arraystretch}{1.05}
\small
\tabcolsep 2pt
\caption{\textbf{Detection results on the \nusc Dataset.}
We report \APBEV\,/\,\AP at IoU=$0.5$ for mobile objects under various ranges. Please refer to \autoref{tbl:main} for naming.
This corresponds to the counterpart \autoref{tbl:nusc_025} in the main paper.
\label{tbl:nusc_05}}
\resizebox{.47\textwidth}{!}{%
\begin{tabular}{=l|+C{37pt}|+C{37pt}|+C{37pt}||+C{37pt}}
\multicolumn{1}{c|}{\multirow{2}{*}{Method}} & \multicolumn{4}{c}{\APBEV\,/\,\AP @IoU = 0.5}\\
\cline{2-5}\
 & 0-30 & 30-50 & 50-80 & 0-80 \\\hline
 \ourmethod-PP(R0)
 & \phantom{0}0.0\,/\,\phantom{0}0.0 & \phantom{0}0.0\,/\,\phantom{0}0.0 & \phantom{0}0.0\,/\,\phantom{0}0.0 & \phantom{0}0.0\,/\,\phantom{0}0.0\\
 
 \ourmethod-PP(R10) 
 & - & - & - & -\\
 
\ourmethod(R0) & \phantom{0}8.4\,/\,\phantom{0}2.9 & \phantom{0}0.3\,/\,\phantom{0}0.1 & \phantom{0}0.1\,/\,\phantom{0}0.0 & \phantom{0}3.0\,/\,\phantom{0}0.9\\

 \ourmethod(R10) & 11.0\,/\,\phantom{0}7.6 & \phantom{0}0.4\,/\,\phantom{0}0.0 & \phantom{0}0.0\,/\,\phantom{0}0.0 & \phantom{0}3.9\,/\,\phantom{0}2.2 \\\hline\hline
 
 \rowstyle{\leavevmode\color{gray}} Sup. (\nusc) & 29.5\,/\,26.3 & \phantom{0}8.4\,/\,\phantom{0}6.1 & \phantom{0}2.4\,/\,\phantom{0}1.1 & 15.5\,/\,13.3 \\
 \hline
\end{tabular}%
}
\vspace{-10pt}
\end{table}

\begin{table}[!t]
\centering
\renewcommand{\arraystretch}{1.05}
\small
\tabcolsep 2pt
\caption{\textbf{Detection performance on the \kitti validation set with models trained on the \lyft dataset.}
We report \APBEV\,/\,\AP with IoU=$0.5$ for mobile objects under various ranges. Please refer to \autoref{tbl:main} for naming.
This corresponds to the counterpart \autoref{tbl:kitti_025} in the main paper.
\label{tbl:kitti_05}}
\resizebox{.47\textwidth}{!}{%
\begin{tabular}{=l|+C{36pt}|+C{36pt}|+C{36pt}||+C{36pt}}
\multicolumn{1}{c|}{\multirow{2}{*}{Method}} & \multicolumn{4}{c}{\APBEV\,/\,\AP @ IoU = 0.5}\\
\cline{2-5}\
 & 0-30 & 30-50 & 50-80 & 0-80 \\\hline
 \ourmethod-PP (R10) 
 & 50.5\,/\,48.5 & 10.3\,/\,\phantom{0}8.7 & \phantom{0}0.2\,/\,\phantom{0}0.1 & 35.6\,/\,33.8\\
 \ourmethod (R10)
 & 62.1\,/\,57.6 & 41.7\,/\,32.3 & \phantom{0}5.3\,/\,\phantom{0}2.0 & 51.1\,/\,46.0\\
 \ourmethod (R40)
&57.7\,/\,52.7 & 44.1\,/\,40.1 & 11.6\,/\,7.2 & 49.3\,/\,44.6\\
 \hline\hline
 \rowstyle{\leavevmode\color{gray}} Sup. (\lyft) & 79.0\,/\,76.7 & 47.7\,/\,42.8 & 19.0\,/\,12.4 & 65.3\,/\,62.5 \\
 \rowstyle{\leavevmode\color{gray}} Sup. (\kitti) & 85.4\,/\,83.3 & 69.5\,/\,66.9 & 41.2\,/\,35.0 & 78.3\,/\,76.0 \\
 \hline
\end{tabular}%
}
\vspace{-10pt}
\end{table}

\begin{table}[!t]
\centering
\renewcommand{\arraystretch}{1.05}
\small
\tabcolsep 2pt
\caption{\textbf{The precision and recall of the ``labels'' on the \lyft dataset ``train'' split.} 
We report the \textbf{\emph{precision\,/\,recall}} rate with BEV IoU=$0.5$ for mobile objects under various ranges. Please refer to \autoref{tbl:main} for naming.
This corresponds to the counterpart \autoref{tbl:label_quality} in the main paper.
\label{tbl:label_quality_iou05}}
\vspace{-8pt}
\resizebox{.47\textwidth}{!}{%
\begin{tabular}{=l|+C{37pt}|+C{37pt}|+C{37pt}||+C{37pt}}
\multicolumn{1}{c|}{\multirow{2}{*}{Method}} & \multicolumn{4}{c}{Precision\,/\,Recall @ IoU = 0.5} 
\\\cline{2-5}
 & 0-30 & 30-50 & 50-80 & 0-80 
 \\\hline
 \ourmethod-PP (seed)  
 & 44.4\,/\,50.3 & \phantom{0}8.5\,/\,16.1 & \phantom{0}2.3\,/\,\phantom{0}2.6 & 16.3\,/\,22.7
 \\
 \ourmethod (seed)
 & 55.8\,/\,43.5 & 28.8\,/\,19.3 & 15.6\,/\,\phantom{0}4.4 & 38.9\,/\,22.2
 \\
 \ourmethod (R0)
 & 79.4\,/\,55.9 & 51.3\,/\,34.0 & 30.0\,/\,\phantom{0}9.6 & 59.3\,/\,33.0 
 \\
 \ourmethod (R10)
 & 82.2\,/\,63.4 & 65.9\,/\,55.0 & 38.7\,/\,28.9 & 62.9\,/\,49.3 
 \\
   \ourmethod (R40)
 & 83.1\,/\,66.2 & 77.6\,/\,67.0 & 69.0\,/\,55.3 & 77.2\,/\,63.3 
 \\
 \hline
\end{tabular}%
}
\vspace{-8pt}
\end{table}

\begin{table}[!t]
\centering
\renewcommand{\arraystretch}{1.05}
\small
\tabcolsep 2pt
\caption{\textbf{Common sense vs self-training.} 
We report \APBEV\,/\,\AP with IoU=$0.5$ for mobile objects under various ranges. Seed and ST column mean how much data are used as seed data and self-training data respectively; FT stands for filtering by \pp score during self-training.
This corresponds to the counterpart \autoref{tbl:ablation} in the main paper.
\label{tbl:ablation_05}}
\vspace{-8pt}
\resizebox{.47\textwidth}{!}{%
\begin{tabular}{=C{21pt}|C{21pt}|C{10pt}|+C{38pt}|+C{38pt}|+C{38pt}||+C{38pt}} 
\multicolumn{3}{c|}{\multirow{1}{*}{Combinations}} & \multicolumn{4}{c}{\APBEV\,/\,\AP @ IoU = 0.5}\\
\hline
 Seed & ST & FT & 0-30 & 30-50 & 50-80 & 0-80 
 \\\hline
 5\% & 5\% & 
 & 43.9\,/\,40.3 & 35.6\,/\,30.8 & \phantom{0}7.8\,/\,\phantom{0}5.1 & 30.2\,/\,26.2
 \\
 5\% & 5\% & \checkmark 
 & 47.7\,/\,40.0 & 40.8\,/\,37.7 & 10.1\,/\,\phantom{0}8.6 & 33.9\,/\,29.8
 \\
 5\% & 100\% & 
 & 55.2\,/\,51.1 & 46.4\,/\,38.0 & 13.8\,/\,\phantom{0}8.4 & 40.2\,/\,34.2
 \\
 100\% & 5\% & 
 & 44.6\,/\,40.4 & 38.7\,/\,33.0 & 16.3\,/\,\phantom{0}7.2 & 34.4\,/\,28.0
\\
 100\% & 5\% & \checkmark 
 & 48.5\,/\,43.0 & 43.4\,/\,35.6 & 18.2\,/\,\phantom{0}8.3 & 38.2\,/\,30.2
 \\
 100\% & 100\% & 
 & 57.8\,/\,52.2 & 46.4\,/\,38.4 & 14.4\,/\,\phantom{0}8.6 & 41.4\,/\,34.3
 \\
 100\% & 100\% & \checkmark 
 & 56.8\,/\,51.3 & 51.4\,/\,40.5 & 19.2\,/\,\phantom{0}9.0 & 44.1\,/\,35.5 
 \\
 \hline
\end{tabular}%
}
\vspace{-10pt}
\end{table}

\section{Detailed evaluation by object types}
In \autoref{tbl:recall_by_class}, we include detailed evaluations (BEV IoU$=0.5$, BEV IoU$=0.25$ and by different depth ranges) of the recall of different object types in the \lyft test set. This corresponds to \autoref{tbl:recall_by_class_iou025_summary} in the main paper.

\section{Corresponding IoU=0.5 results}
We list the IoU=0.5 counterparts of \Cref{tbl:main,tbl:ablation,tbl:nusc_025,tbl:label_quality,tbl:kitti_025,tbl:recall_by_class_iou025_summary} in \Cref{tbl:main_iou05,tbl:nusc_05,tbl:kitti_05,tbl:label_quality_iou05,tbl:ablation_05}.

\begin{figure}[!t]
    \centering
    \begin{subfigure}{0.48\linewidth}
    \centering
    \includegraphics[width=\linewidth]{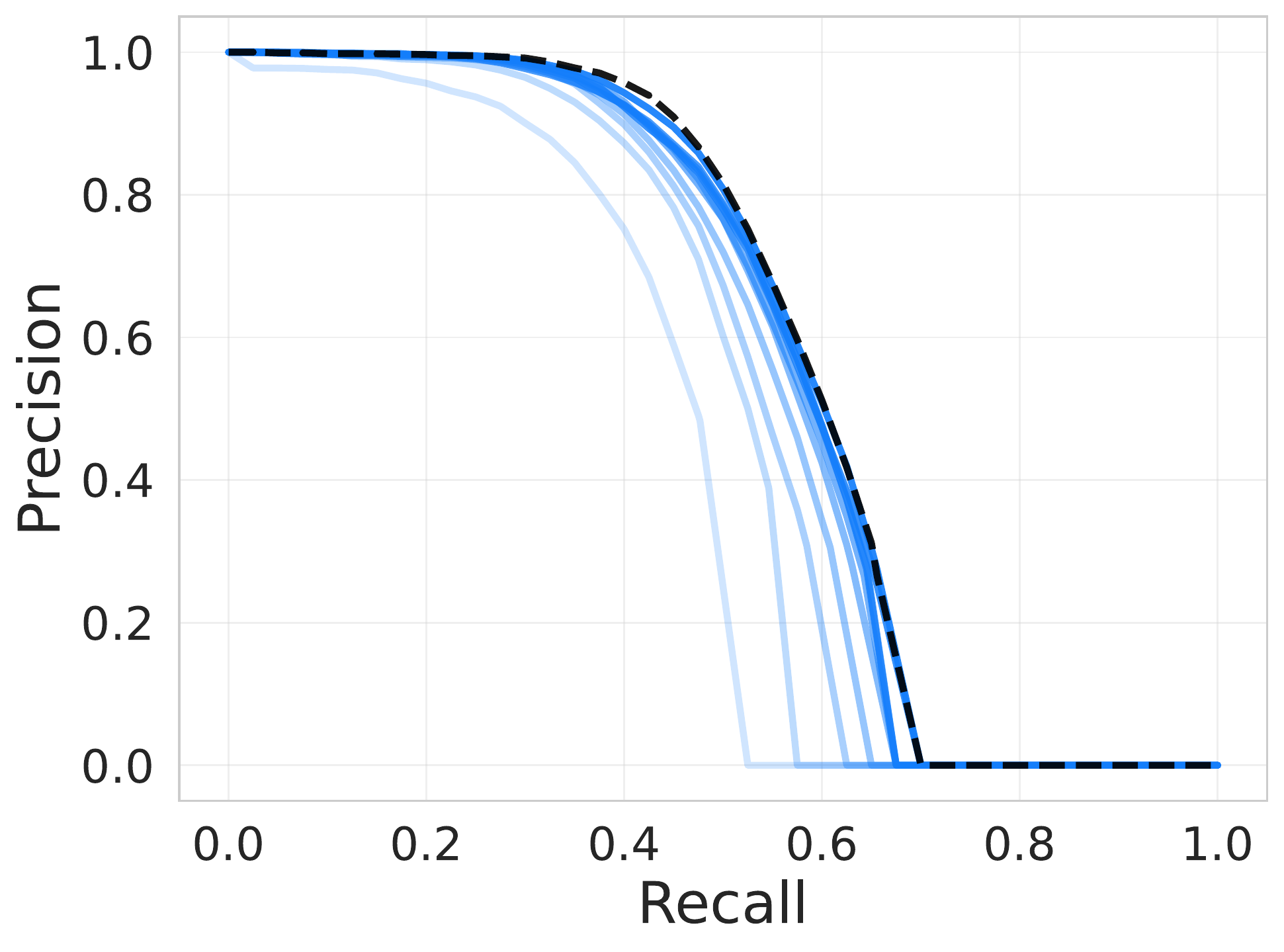}
    \caption{IoU=0.25 \label{fig:ablation_pr_bev025}}
    \end{subfigure}
    \begin{subfigure}{0.48\linewidth}
    \centering
    \includegraphics[width=\linewidth]{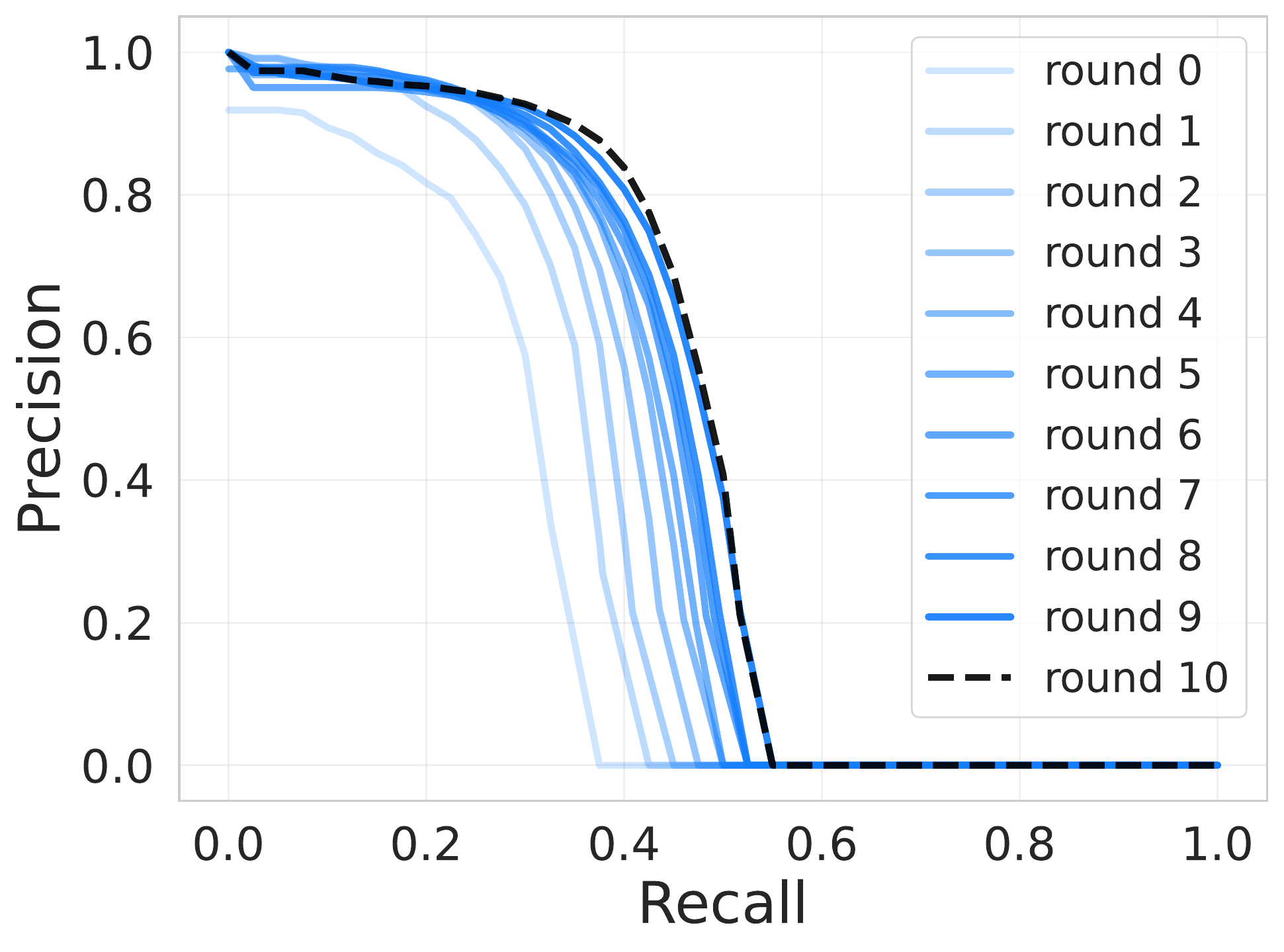}
    \caption{IoU=0.5 \label{fig:ablation_pr_bev05}}
    \end{subfigure}
    \vspace{-8pt}
    \caption{\textbf{Number of self-training rounds \vs precision-recall curves.} We show the precision-recall curves under \APBEV with IoU=0.5 and IoU=0.25 for mobile objects in 0-80\,m on \lyft test set from models trained with different rounds of self-training.
    \label{fig:ablation_pr}
    }
\end{figure}
\section{Precision-recall evaluation}
In \autoref{fig:ablation_pr}, we show how PR curve changes with different rounds of self-training: the max recall improves gradually while keeping high precision.
This aligns with the expanded recall of the training set described above, and with what we observe qualitatively in \autoref{fig:teaser}.

\section{More qualitative results}
\begin{figure*}[htb]
    \centering
    \includegraphics[width=\linewidth]{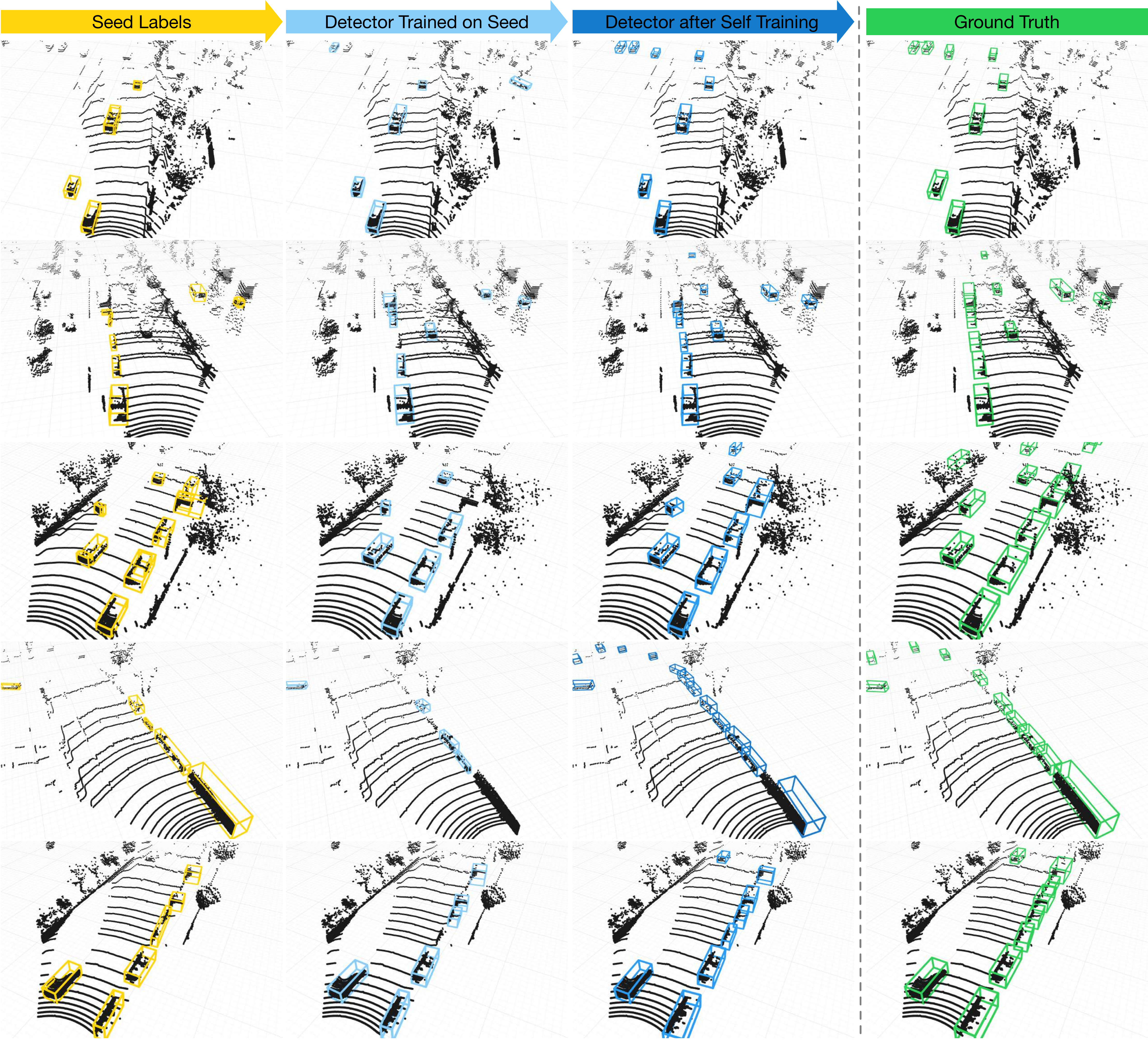}
    \vspace{-15pt}
    \caption{\textbf{Visualizations of \ourmethod outputs.} We show additional visualizations of \lidar scans from three scenes in the \lyft dataset. From left to right: seed labels, detections trained on seed, detections after self training, and ground truth bounding boxes.
    \label{fig:self-training}
    }
\end{figure*}

We show visualizations for additional qualitative results in \autoref{fig:self-training} for 5 additional \lidar scenes. Visualizations show the progression of \ourmethod from seed generation, to detector trained on seed label set, to detector after 10 rounds of self training, and finally the ground truth bounding boxes. Observe that the detections obtain higher recall and learns a correct prior over object shapes as the method progresses.

\end{document}